\documentclass[11pt,a4paper,oneside]{amsart}
\usepackage[colorlinks,citecolor=blue,urlcolor=blue]{hyperref}
\usepackage[foot]{amsaddr}

\usepackage[
hmarginratio={1:1},     
vmarginratio={1:1},     
textwidth=450pt,        
textheight=700pt,
heightrounded,          
]{geometry}
\usepackage{graphicx}
\usepackage{amsmath}
\usepackage{amsfonts}
\usepackage{natbib}
\usepackage{subcaption,graphicx}
\usepackage{amssymb}

\usepackage{bm}
\usepackage{bbm}

\usepackage{etoolbox}

\makeatletter
\patchcmd{\@makecaption}
{\parbox}
{\advance\@tempdima-\fontdimen2} 
{}{}
\makeatother

\usepackage{xargs}                      
\usepackage[colorinlistoftodos,prependcaption,textsize=tiny,textwidth=2.5cm]{todonotes}
\usepackage{marginnote}

\usepackage{enumitem}   
\usepackage{amsmath}
\usepackage{amsthm}
\DeclareMathOperator{\sign}{sign}

\newcommand{\notee}[2][]{{%
		\let\marginpar\marginnote
		\reversemarginpar
		\renewcommand{\baselinestretch}{0.8}%
		\todo[linecolor=red,backgroundcolor=red!25,bordercolor=red,#1]{#2}}}

\newcommand{\rulesep}{\unskip\ \vrule\ }

\theoremstyle{remark}

\title[Prediction explanation with dependence]{Explaining individual predictions when features are dependent:\\ More accurate approximations to Shapley values}

\author{Kjersti Aas\textsuperscript{1}}
\address{\textsuperscript{1,2,3}Norwegian Computing Center, P.O. Box 114, Blindern, N-0314 Oslo, Norway.}
\email{kjersti.aas@nr.no}

\author{Martin Jullum\textsuperscript{2}}
\email{martin.jullum@nr.no}
\author{Anders L{\o}land\textsuperscript{3}}
\email{anders.loland@nr.no}

\begin{document}

\begin{abstract}
Explaining complex or seemingly simple machine learning models is an important practical problem.
We want to explain individual predictions from a complex machine learning model by learning simple, interpretable explanations. Shapley values is a game theoretic concept that can be used for this purpose.
The Shapley value framework has a series of desirable theoretical properties, and can in principle handle any predictive model. Kernel SHAP is a  computationally efficient approximation to Shapley values
in higher dimensions. Like several other existing methods, this approach assumes that the features are independent, which may give very wrong explanations. This is the case even if a simple linear model is used for predictions.
In this paper, we extend  the Kernel SHAP method to handle dependent features. We provide several examples of linear and non-linear models with various degrees of feature dependence, where our method gives more accurate
approximations to the true Shapley values. We also propose a method for aggregating individual Shapley values, such that the prediction can be explained by groups of dependent variables.
\end{abstract}

\maketitle

\section{Introduction}\label{sec:intro}

Interpretability is crucial when a complex machine learning model is to be applied in areas such as medicine \citep{obermeyer16}, fraud detection \citep{sudjianto10} or credit scoring \citep{kvamme2018}. In many applications, complex hard-to-interpret machine learning models like deep neural networks, random forests and gradient boosting machines, are currently outperforming the traditional, and to some extent interpretable, linear/logistic regression models. However, often there is a clear trade-off between model complexity and model interpretability, meaning that it is often hard to understand why these sophisticated models perform so well. This lack of explanation constitutes a practical  issue -- can I trust the  model? \citep{Ribeiro}, 
and a legal 
issue---those who develop the model can be required by law to explain what the model does to those who are exposed to automated decisions -- GDPR \citep{gdpr16}. In response, a new line of research has emerged that focuses on helping users to interpret the predictions from advanced machine learning methods.

Existing work on explaining complex models may be divided into two main categories; global and local explanations. The former tries to describe the model as whole, in terms of which variables/features influenced the general model the most. Two common methods for such an overall explanation is some kind of permutation feature importance or partial dependence plots \citep{Molnar}.
Local explanations, on the other hand, tries to identify how the different input variables/features influenced a specific prediction/output from the model, and are often referred to as individual prediction explanation methods. Such explanations are particularly useful for complex models which behave rather different for different feature combinations, meaning that the global explanation is not representative for the local behavior. 

Local explanation methods may further be divided into two categories: model-specific and model-agnostic (general) explanation methods. In this paper the focus is on the latter. The methods in this 
category usually try to explain individual predictions by learning
simple, interpretable explanations of the model specifically for a given
prediction. Three examples are Explanation Vectors \citep{Baehrens}, 
LIME (Local Interpretable Model-agnostic Explanations \citep{Ribeiro} and Shapley values \citep{Strumbelj,Strumbelj2, Lundberg}. The latter approach, which builds on concepts from  cooperative game theory \citep{Shapley53},
has a series of desirable theoretical properties \citep{Lundberg}.

The Shapley value is a method originally invented for assigning payouts to players depending on their contribution towards the total payout. In the explanation setting, 
the features are the players and the prediction is the total payout. In this framework, the difference between the prediction and the average prediction is perfectly distributed among the features. This property distinguishes Shapley values from other methods like
for example LIME, which does not guarantee perfectly distributed effects. 
It should be noted that LIME
and the Shapley values actually explain two different things. For instance, if the prediction to be explained is the probability of person A crashing his car, the sum of the Shapley values for all features is equal to the difference between this prediction and the mean probability
of a person crashing his car, where the mean is taken over all persons having a driver license. The sum of the LIME values is also equal to the difference between this prediction and a mean probability, but here the mean is taken over all persons ``similar to'' person A. 
That is, Shapley values explain the difference between the prediction and the global average prediction, while LIME explains the difference between the prediction and a local average prediction. Appropriate model explanations should be consistent with how humans 
understand that model. In their study, \cite{Lundberg} found a much stronger agreement between human explanations and Shapley values than with LIME. 

Shapley values have also been used for measuring global feature importance. For instance, it has been used to partition the $R^2$ quantity among the $d$ features in a linear regression model 
(``Shapley regression values''), both assuming independent features \citep{Owen14}, and more recently also for dependent features \citep{Song16, Owen17}. 

The main disadvantage of the Shapley value is that the computational
time grows exponentially and becomes intractable for more than say ten
features. This has led to approximations like the Shapley Sampling
Values \citep{Strumbelj,Strumbelj2} and Kernel SHAP
\citep{Lundberg}. The latter requires less computational power to
obtain a similar approximation accuracy. Hence, in this paper, the
focus is on the Kernel SHAP method. While having many desirable
properties, this method assumes feature independence. 
In observational studies and machine learning problems, it is very rare that the features are statistically independent.
If there is high degree of dependence/correlation among some or all the features, the resulting explanations might be very wrong. This is the case even if a simple linear model is used. 
Recently, \cite{Lundberg2} have proposed a method based on Shapley Values denoted Tree SHAP, which is said to assume ``less'' feature independence in the sense that it
accounts for some of the dependence, but not all. As we be apparent from our simulations experiments, this method does not deliver in terms of accuracy, being potentially highly inaccurate when the features are dependent.  Moreover, it is not universally applicable, but specially designed for tree ensemble methods like XGBoost \citep{Chen2016}.

In this paper, we extend the Kernel SHAP method to handle dependent features, 
using some of the ideas from the latter. To the best of our knowledge, there has been no previous research on what dependency between the features means for the Shapley values in the explanation setting, nor a suggestion on how 
to handle such dependence. Hence, we regard this paper to be the first to account for dependence within Shapley value based individual prediction explanation. Moreover, we propose a method for clustering Shapley values corresponding to dependent features, 
improving the presentation of feature contribution for individual predictions in the presence of feature dependence. Our methodology has been implemented in an \verb,R,-package currently available at: \url{https://github.com/NorskRegnesentral/shapr}.

The rest of this paper is organized as follows. Section \ref{models} reviews the Shapley values and the Kernel SHAP method. In Section \ref{dependence} we describe our proposed approaches for 
capturing dependence, while Section \ref{experiment} gives the results
from several experiments validating these methods. An approach for clustering of
Shapley values corresponding to feature dependence is described in
Section \ref{sec:clustering} and, finally, Section \ref{summary} contains some 
concluding remarks.

\section{The exact Shapley value and the Kernel SHAP approximation} \label{models}

In this section we first give the definition of the Shapley Value from game theory in Section \ref{ExactShap}, explain its use in the context of explaining individual predictions in Section \ref{PredExplain}, and then we describe the Kernel SHAP method in Section \ref{kernelShap}.

\subsection{The exact Shapley value and cooperative game theory}\label{ExactShap}

Consider a cooperative game with $M$ players aiming at maximizing a payoff, and let $S \subseteq \mathcal{M} = \{ 1,\ldots,M\}$ be a subset consisting of $|\mathcal{S}|$ players.
Assume that we have a ``contribution'' function $v(\mathcal{S})$ that maps subsets of players to the real numbers, called the worth or contribution of coalition $\mathcal{S}$. It describes the total expected sum of payoffs 
the members of $\mathcal{S}$ can obtain by cooperation. The Shapley value \citep{Shapley53} is one way to distribute the total gains to the players, assuming that they all collaborate. 
It is a "fair" distribution in the sense that it is the only distribution with certain desirable properties listed below. According to the Shapley value, the amount that player $j$ gets is
\begin{equation}\label{shapleyValue}
\phi_j(v) = \phi_j = \sum_{\mathcal{S} \subseteq \mathcal{M} \setminus\{j\}} \frac{|\mathcal{S}| ! (M-| \mathcal{S}| - 1)!}{M!}(v(\mathcal{S}\cup \{j\})-v(\mathcal{S})), \quad j=1,\ldots,M,
\end{equation}
that is, a weighted mean over all subsets $\mathcal{S}$ of players not containing player $j$. Note that the empty set $\mathcal{S}=\emptyset$ is also part of this sum.
The formula can be interpreted as follows: Imagine the coalition being formed for one player at a time, 
with each player demanding their contribution $v(\mathcal{S} \cup \{j\}) - v(\mathcal{S})$ as a fair compensation. Then, for each player, compute the average of this contribution over all possible combinations 
in which the coalition can be formed, yielding a weighted mean.

To illustrate the application of \eqref{shapleyValue}, let us consider a game with three players such that $\mathcal{M} = \{1,2,3\}$. Then, there are 8 possible subsets; $\emptyset, \{1\}, \{2\}, \{3\}, \{1,2\}, \{1,3\}, \{2,3\}$, 
and $\{1,2,3\}$.  Using \eqref{shapleyValue}, the Shapley values for the three players are given by
\begin{eqnarray*}
	\scriptstyle{\phi_{1}} &=& \scriptstyle{\frac{1}{3}\,\big(v(\{1,2,3\})-v(\{2,3\})\big)+ \frac{1}{6}\,\big(v(\{1,2\})-v(\{2\})\big)+\frac{1}{6}\,\big(v(\{1,3\})-v(\{3\})\big)+\frac{1}{3}\,\big(v(\{1\})-v(\emptyset)\big)},\\
	\scriptstyle{\phi_{2}} &=& \scriptstyle{\frac{1}{3}\,\big(v(\{1,2,3\})-v(\{1,3\})\big)+ \frac{1}{6}\,\big(v(\{1,2\})-v(\{1\})\big)+\frac{1}{6}\,\big(v(\{2,3\})-v(\{3\})\big)+\frac{1}{3}\,\big(v(\{2\})-v(\emptyset)\big)},\\
	\scriptstyle{\phi_{3}} &=& \scriptstyle{\frac{1}{3}\,\big(v(\{1,2,3\})-v(\{1,2\})\big)+ \frac{1}{6}\,\big(v(\{1,3\})-v(\{1\})\big)+\frac{1}{6}\,\big(v(\{2,3\})-v(\{2\})\big)+\frac{1}{3}\,\big(v(\{3\})-v(\emptyset)\big)}.
\end{eqnarray*}
Let us also define the non-distributed gain $\phi_{0}   = v(\emptyset)$, that is, the fixed payoff which is not associated to the actions of any of the players, although this is often zero for coalition games.

By summarizing the right hand sides above, we easily see that they add
up to the total worth of the game:
$\phi_{0} + \phi_{1}+\phi_{2}+\phi_{3} = v(\{1,2,3\})$.

The Shapley value has the following desirable properties
\begin{description}
	\item[Efficiency:] The total gain is distributed:
	\[ \sum_{j = 0}^M \phi_j = v(\mathcal{M})\]
	\item[Symmetry:] If $i$ and $j$ are two players who contribute equally to all possible coalitions, i,e.
	\[v(\mathcal{S}\cup \{i\})=v(\mathcal{S}\cup \{j\})\]
	for every subset $\mathcal{S}$ which contains neither $i$ nor $j$, then their Shapley values are identical:
	\[\phi_{i}=\phi_{j}.\]
	\item[Dummy player:] If $v(\mathcal{S}\cup \{j\})=v(\mathcal{S})$ for a player $j$ and all coalitions $\mathcal{S}\subseteq \mathcal{M}\setminus\{j\}$, then $\phi_j=0$.
	\item[Linearity:] If two coalition games described by gain functions $v$ and $w$ are combined, then the distributed gains correspond to the gains derived from $v$ and the gains derived from $w$: 
	\[\phi_{i}(v+w)=\phi_{i}(v)+\phi_i(w),\]
	for every $i$. Also, for any real number $a$ we have that 
	\[ \phi_{i}(a\,v)= a\phi_i(v).\]
\end{description}
The Shapley values is the only set of values satisfying the above properties, see \cite{Shapley53} and \cite{Young} for proofs.

\subsection{Shapley values for prediction explanation}\label{PredExplain}

Consider a classical machine learning scenario where a training set $\{y^i,\boldsymbol{x}^i\}_{i=1,\ldots,n_{\text{train}}}$ of size $n_{\text{train}}$ has been used to train a predictive model $f(\boldsymbol{x})$ attempting to resemble a response value $y$ as closely 
as possible. Assume now that we want to explain the prediction from the model  $f(\boldsymbol{x}^*)$, for a specific feature vector $\boldsymbol{x}=\boldsymbol{x}^*$.
\citet{Strumbelj,Strumbelj2} and \citet{Lundberg} suggest to do this using Shapley values. By moving from game theory to decomposing an individual prediction into feature contributions, the single prediction takes the place of the payout, and the features take 
the place of the players. 
We have that the prediction $f(\boldsymbol{x}^*)$ is decomposed as follows 
\[ f(\boldsymbol{x}^*) = \phi_0 + \sum_{j=1}^M \phi_j^*,\] 
where $\phi_0=\mbox{E}[f(\boldsymbol{x})]$ and $\phi_j^*$ is the $\phi_j$ for the prediction $\boldsymbol{x}=\boldsymbol{x}^*$. That is, the Shapley values explain the difference between the prediction $y^* = f(\boldsymbol{x}^*)$ and the global average prediction. 
A model of this form is an additive feature attribution method, and 
it is the only additive feature attribution method that adhers to the properties listed in Section \ref{ExactShap} \citep{Lundberg}. In Appendix \ref{properties} we discuss why these properties are useful in the prediction explanation setting. Note that LIME, another 
well-known additive feature attribution method, does not satisfy the four properties, which may lead to inconsistent explanations.

To be able to compute the Shapley values in the prediction explanation setting, we need to define the contribution function $v(\mathcal{S})$ for a certain subset $\mathcal{S}$. This function should resemble the value of $f(\boldsymbol{x}^*)$ when we only know the value of 
the subset $\mathcal{S}$ of these features. To quantify this, we follow \cite{Lundberg} and use the expected output of the predictive model, conditional on the feature values $\boldsymbol{x}_{\mathcal{S}}=\boldsymbol{x}_{\mathcal{S}}^*$ of this subset:
\begin{equation}\label{contrib.func}
v(\mathcal{S}) = \mbox{E}[f(\boldsymbol{x})|\boldsymbol{x}_{\mathcal{S}}=\boldsymbol{x}_{\mathcal{S}}^*].
\end{equation}
Other measures, such as the conditional median, may also be appropriate. However, the conditional expectation summarises the whole probability distribution and it is the most common estimator in prediction applications. When viewed as a 
prediction for $f(\boldsymbol{x}^*)$, it is also the minimiser of the commonly used squared error loss function.

If the predictive model is a linear regression model $f(\boldsymbol{x}) = \beta_0 + \sum_{j=1}^M\beta_j\,x_{j}$, where all features $x_j, j=1,\ldots,M$ are independent, we show in Appendix \ref{appLinear} that under \eqref{contrib.func}, the Shapley values take the simple form:
\begin{equation}\label{linform}
\phi_{0} = \beta_0 + \sum_{j=1}^M \beta_jE[x_j] , \quad \text{and} \quad \phi_{j} = \beta_j\,(x^*_{j}-E[x_j]), \quad j = 1,\ldots,M.
\end{equation}
Note, that for ease of notation, we have here and in the rest of the paper dropped the superscript * for the $\phi_{j}$ values. Every prediction $f(\boldsymbol{x}^*)$ to be explained will result in different sets of $\phi_{j}$ values.

To the best of our knowledge, no explicit formula like \eqref{linform} exists for the general case of dependent features with non-linear models.  With $M$ features, the number of possible
subsets involved in \eqref{shapleyValue} is $2^M$. Hence, the number of possible subsets increases exponentially when $M$ increases, meaning that the exact solution to this problem
becomes computationally intractable when we have more than a few features. In Section \ref{kernelShap}, we shall see how a clever approximation method may be used to (partly) overcome this issue.


In addition to a computationally tractable approximation for computing the Shapley values, applying the above method in practice requires an estimate of the expectation in \eqref{contrib.func} for all $\boldsymbol{x}_{\mathcal{S}}$.
The main methodological contribution of this paper is describing, developing, and comparing methodology to appropriately estimate these expectations. In Section \ref{expectation1} we describe the state-of-art method for
determining the expectations before we describe our proposed approaches in Section \ref{dependence}.


\subsection{Kernel SHAP}\label{kernelShap}

The Kernel SHAP method of \cite{Lundberg} aims at estimating Shapley values under \eqref{contrib.func} in practical situations. The method may be divided in two separate parts: 
\begin{enumerate}[label=(\roman*)]
	\item A clever computationally tractable approximation for computing the Shapley values of \eqref{shapleyValue}
	\item A simple method for estimating $v(\mathcal{S})$
\end{enumerate}
In \cite{Lundberg}, the method is presented in a somewhat limited form without full detail. In order to facilitate the understanding of the method and our consecutive improvements to the method, we shall therefore in this section carefully re-state the Kernel SHAP method. We will 
first present part i), assuming $v(\mathcal{S})$ is known, and then present the method for estimating $v(\mathcal{S})$ used by Kernel SHAP.


%
%
%
%
%
\subsubsection{Approximated weighted least squares}

There are several alternative equivalent formulas for the Shapley values. \cite{Charnes1988} and later \cite{Lundberg} define the Shapley values as the optimal solution of a certain weighted least squares (WLS) problem.

In its simplest form, the WLS problem can be stated as the problem of minimizing
\begin{equation}
\sum_{\mathcal{S}\subseteq \mathcal{M}} (v(\mathcal{S}) - (\phi_0 + \sum_{j\in \mathcal{S}}\phi_j))^2 k(M,\mathcal{S}), \label{eq:KS.problem}
\end{equation}
with respect to $\phi_0,\ldots,\phi_M$, where $k(M,\mathcal{S}) = (M-1)/(\tbinom{M}{|\mathcal{S}|}\,|\mathcal{S}|\,(M-|\mathcal{S}|))$, are denoted the Shapley kernel weights. Let us write $\boldsymbol{Z}$ for the $2^M \times (M+1)$ binary matrix representing all possible combinations of inclusion/exclusion of the $M$ features, where the first column is 1 for every row, while entry $j+1$ of row $l$ is 1 if feature $j$ is included in combination $l$, and 0 otherwise. Let also 
$\boldsymbol{v}$ be a vector containing $v(\mathcal{S})$, and $\boldsymbol{W}$ be the $2^{M} \times 2^{M}$ diagonal matrix containing $k(M,|\mathcal{S}|)$, where $\mathcal{S}$ in both cases resembles the feature combinations of the corresponding row in $\boldsymbol{Z}$. Then \eqref{eq:KS.problem} may be rewritten to 
\begin{equation}
(\boldsymbol{v}-\boldsymbol{Z}\boldsymbol{\phi})^T \boldsymbol{W} (\boldsymbol{v}-\boldsymbol{Z}\boldsymbol{\phi}) \label{eq:KS.problem.matrix},
\end{equation}
for which the solution is
\begin{equation}
\mathbf{\phi} = \left(\boldsymbol{Z}^T\boldsymbol{W}\boldsymbol{Z}\right)^{-1}\boldsymbol{Z}^T\boldsymbol{W}\boldsymbol{v}. \label{eq:KS.formula}
\end{equation}
In practice, the infinite Shapley kernel weights $k(M,M)=k(M,0) = \infty$ may be set to a large constant $C$, for example $C=10^6$, or imposing the constraints that $\phi_0=v(\emptyset)$ and $\sum_{j=0}^M \phi_j= v(\mathcal{M})$ into the problem. 


When the model contains more than a few features $M$, computing the right hand side of \eqref{eq:KS.formula} is still computationally expensive. However, the trick is to use the weighted least squares formulation to approximate \eqref{eq:KS.formula}. The Shapley kernel weights have very different sizes, meaning that the majority of the subsets $\mathcal{S}$, that is, the rows in $\boldsymbol{Z}$, contributes very little to the Shapley value. Hence, assuming that we have an proper approximation for the elements in $\boldsymbol{v}$, a consistent approximation may be obtained by sampling (with replacement) a subset $\mathcal{D}$ of $\mathcal{M}$ from a probability distribution following the Shapley weighting kernel, and using only those rows $\boldsymbol{Z}_{\mathcal{D}}$ of $\boldsymbol{Z}$ and elements $\boldsymbol{v}_{\mathcal{D}}$ of $\boldsymbol{v}$ in the computation. As the Shapley kernel weights are used in the sampling, the sampled subsets are weighted equally in the new least squares problem.
Note that $\mathcal{S}=\emptyset$ and $\mathcal{S}=\mathcal{M}$ are excluded from the sampling procedure. 
As above, their corresponding $\boldsymbol{Z}$-rows are appended to $\boldsymbol{Z}_{\mathcal{D}}$, their $v(\mathcal{S})$-elements are appended to $\boldsymbol{v}_{\mathcal{D}}$, and the diagonal matrix $\boldsymbol{W}_{\mathcal{D}}$ is extended with two diagonal 
elements equal to $C$.
This procedure gives the following approximation to \eqref{eq:KS.formula}:\footnote{The bulk of information regarding the approximation of the WLS problem was obtained through personal communication with Scott Lundberg.} 
\begin{equation}
\mathbf{\phi} = \left[\left(\boldsymbol{Z}_{\mathcal{D}}^T\boldsymbol{W}_{\mathcal{D}}\boldsymbol{Z}_{\mathcal{D}}\right)^{-1}\boldsymbol{Z}_{\mathcal{D}}^T\boldsymbol{W}_{\mathcal{D}}\right]\boldsymbol{v}_{\mathcal{D}} = \boldsymbol{R}_{\mathcal{D}}\boldsymbol{v}_{\mathcal{D}}. \label{eq:KS.formula.approx}
\end{equation}
A practical consequence of using \eqref{eq:KS.formula.approx} as opposed to \eqref{shapleyValue} is that when explaining several predictions, which typically is the case, the matrix operations producing the $(M+1) \times |\mathcal{D}|$ matrix $\boldsymbol{R}_{\mathcal{D}}$ only has to be carried out once. All that is needed to explain different predictions from the same model (provided $\boldsymbol{v}_{\mathcal{D}}$ is pre-computed) is to perform the matrix multiplication of $\boldsymbol{R}_{\mathcal{D}}$ and $\boldsymbol{v}_{\mathcal{D}}$ for the different $\boldsymbol{v}_{\mathcal{D}}$.

\subsubsection{Estimating the contribution function under feature independence} \label{expectation1}

When computing the vector $\boldsymbol{v}$, we need the $v(\mathcal{S})$ values for all possible feature subsets represented by the matrix $\boldsymbol{Z}$. As stated in Section \ref{ExactShap}, 
the contribution value $v(\mathcal{S})$ for a certain subset $\mathcal{S}$ is defined as 
\[ v(\mathcal{S}) = \mbox{E}[f(\boldsymbol{x})|\boldsymbol{x}_{\mathcal{S}}=\boldsymbol{x}_{\mathcal{S}}^*].\]
Let $\bar{\mathcal{S}}$ denote the complement of $\mathcal{S}$, such that $\boldsymbol{x}_{\bar{\mathcal{S}}}$ is the part of $\boldsymbol{x}$ not in $\boldsymbol{x}_{\mathcal{S}}$. Then, the expected value may be computed as follows
\begin{align}
E[f(\boldsymbol{x})|\boldsymbol{x}_{\mathcal{S}}=\boldsymbol{x}_{\mathcal{S}}^*] &= E[f(\boldsymbol{x}_{\bar{\mathcal{S}}},\boldsymbol{x}_{\mathcal{S}})|\boldsymbol{x}_{\mathcal{S}}=\boldsymbol{x}_{\mathcal{S}}^*] 
= \int f(\boldsymbol{x}_{\bar{\mathcal{S}}},\boldsymbol{x}_{\mathcal{S}}^*)\,p(\boldsymbol{x}_{\bar{\mathcal{S}}}|\boldsymbol{x}_{\mathcal{S}}=\boldsymbol{x}_{\mathcal{S}}^*)d\boldsymbol{x}_{\bar{\mathcal{S}}}, \label{eq:int}
\end{align}
where $p(\boldsymbol{x}_{\bar{\mathcal{S}}}|\boldsymbol{x}_{\mathcal{S}}=\boldsymbol{x}_{\mathcal{S}}^*)$ is the conditional distribution of $\boldsymbol{x}_{\bar{\mathcal{S}}}$ given $\boldsymbol{x}_{\mathcal{S}}=\boldsymbol{x}_{\mathcal{S}}^*$.
Hence, to be able to compute the exact $v(\mathcal{S})$ values we need the conditional distribution $p(\boldsymbol{x}_{\bar{\mathcal{S}}}|\boldsymbol{x}_{\mathcal{S}}=\boldsymbol{x}_{\mathcal{S}}^*)$, which seldom is known. 
In this step of the procedure, the Kernel SHAP method assumes feature independence, replacing $p(\boldsymbol{x}_{\bar{\mathcal{S}}}|\boldsymbol{x}_{\mathcal{S}})$ by $p(\boldsymbol{x}_{\bar{\mathcal{S}}})$. Using the training set, 
that is, the data used to train the model $f(\cdot)$, as the empirical distribution of $\boldsymbol{x}$, the integral in \eqref{eq:int} can be approximated by
\begin{equation}\label{expectationIndep}
v_{\text{KerSHAP}}(\mathcal{S}) = \frac{1}{K}\sum_{k=1}^K f(\boldsymbol{x}_{\bar{\mathcal{S}}}^k,\boldsymbol{x}_{\mathcal{S}}^*),
\end{equation}
where $\boldsymbol{x}_{\bar{\mathcal{S}}}^k, k=1,\ldots,K$ are samples from the training data. Due to the independence assumption, they are sampled independently of $\boldsymbol{x}_{\mathcal{S}}$.

\section{Incorporating dependence into the Kernel SHAP method}\label{dependence}

If the features in a given model are highly dependent, the Kernel SHAP method may give a completely wrong answer. As stated in Section \ref{sec:intro}, it is very rare that features in real datasets are statistically independent. The only place in the Kernel SHAP method 
where the independence assumption $p(\boldsymbol{x}_{\bar{\mathcal{S}}}|\boldsymbol{x}_{\mathcal{S}})= p(\boldsymbol{x}_{\bar{\mathcal{S}}})$ is used, is when approximating the integral in \eqref{eq:int}. Apart from this rough assumption, the Kernel SHAP framework stands out as a 
clever and fruitful way to approximate the Shapley values. It would therefore be desirable to incorporate dependence into the Kernel SHAP method by relaxing the independence assumption. This can be done by estimating/approximating 
$p(\boldsymbol{x}_{\bar{\mathcal{S}}}|\boldsymbol{x}_{\mathcal{S}}=\boldsymbol{x}_{\mathcal{S}}^*)$ directly and 
generate samples from this distribution, instead of generating them independently from $\boldsymbol{x}_{\mathcal{S}}$ as in Section \ref{expectation1}.
We propose four approaches for estimating $p(\boldsymbol{x}_{\bar{\mathcal{S}}}|\boldsymbol{x}_{\mathcal{S}}=\boldsymbol{x}_{\mathcal{S}}^*)$; (i) assuming a Gaussian distribution for $p(\boldsymbol{x})$, (ii) assuming a Gaussian copula distribution for $p(\boldsymbol{x})$,  
(iii) approximating $p(\boldsymbol{x}_{\bar{\mathcal{S}}}|\boldsymbol{x}_{\mathcal{S}}=\boldsymbol{x}_{\mathcal{S}}^*)$ by an empirical (conditional) distribution and (iv) a combination of the empirical approach and either the Gaussian or the Gaussian copula approach. 

\subsection{Multivariate Gaussian distribution}\label{gaussMet}

If we assume that the feature vector $\boldsymbol{x}$ stems from a multivariate Gaussian distribution with some mean vector $\boldsymbol{\mu}$ and covariance matrix $\boldsymbol{\Sigma}$, the conditional distribution $p(\boldsymbol{x}_{\bar{\mathcal{S}}} |\boldsymbol{x}_{\mathcal{S}}=\boldsymbol{x}_{\mathcal{S}}^*)$ is also multivariate Gaussian. In particular, writing $p(\boldsymbol{x}) = p(\boldsymbol{x}_{\mathcal{S}},\boldsymbol{x}_{\bar{\mathcal{S}}})= \text{N}_M(\boldsymbol{\mu},\boldsymbol{\Sigma})$ with $\boldsymbol{\mu}=(\boldsymbol{\mu}_{\mathcal{S}}, \boldsymbol{\mu}_{\bar{\mathcal{S}}})^{\top}$
and 
\[\boldsymbol{\Sigma} = \begin{bmatrix} 
\boldsymbol{\Sigma}_{\mathcal{S}\mathcal{S}} & \boldsymbol{\Sigma}_{\mathcal{S}\bar{\mathcal{S}}} \\
\boldsymbol{\Sigma}_{\bar{\mathcal{S}}\mathcal{S}} & \boldsymbol{\Sigma}_{\bar{\mathcal{S}}\bar{\mathcal{S}}}
\end{bmatrix},
\]
gives $p(\boldsymbol{x}_{\bar{\mathcal{S}}} | \boldsymbol{x}_{\mathcal{S}}=\boldsymbol{x}_{\mathcal{S}}^*) = \text{N}_{|\bar{\mathcal{S}}|}(\boldsymbol{\mu}_{\bar{\mathcal{S}}|\mathcal{S}},\boldsymbol{\Sigma}_{\bar{\mathcal{S}}|\mathcal{S}})$, with 
\begin{equation}\label{mu}
\boldsymbol{\mu}_{\bar{\mathcal{S}}|\mathcal{S}} = \boldsymbol{\mu}_{\bar{\mathcal{S}}} + \boldsymbol{\Sigma}_{\bar{\mathcal{S}}\mathcal{S}}\,\boldsymbol{\Sigma}_{\mathcal{S}\mathcal{S}}^{-1}(\boldsymbol{x}_{\mathcal{S}}^*-\boldsymbol{\mu}_{\mathcal{S}})
\end{equation}
and
\begin{equation}\label{cov}
\boldsymbol{\Sigma}_{\bar{\mathcal{S}}|\mathcal{S}} = \boldsymbol{\Sigma}_{\bar{\mathcal{S}}\bar{\mathcal{S}}}- \boldsymbol{\Sigma}_{\bar{\mathcal{S}}\mathcal{S}}\,\boldsymbol{\Sigma}_{\mathcal{S}\mathcal{S}}^{-1} \boldsymbol{\Sigma}_{\mathcal{S}\bar{\mathcal{S}}}.
\end{equation}        
Hence, instead of sampling from the marginal distribution of $\boldsymbol{x}_{\bar{\mathcal{S}}}$, we can sample from the Gaussian distribution with expectation vector and covariance matrix given by \eqref{mu}
and \eqref{cov}, where the full expectation vector $\boldsymbol{\mu}$ and the full covariance matrix $\boldsymbol{\Sigma}$ are estimated by the sample mean and covariance matrix of the training data, respectively.
Using the samples  $\boldsymbol{x}_{\bar{\mathcal{S}}}^k, k=1,\ldots,K$ from the conditional distribution, the integral in \eqref{eq:int} is finally approximated by \eqref{expectationIndep}.

\subsection{Gaussian copula} \label{gaussCopula}

When the features are far from multivariate Gaussian distributed, one may instead represent the marginals by their empirical distributions, and model the dependence structure by a Gaussian copula.
The definition of a $d$-dimensional copula is a multivariate distribution, $C$, with uniformly distributed marginals U(0,1) on [0,1]. Sklar's theorem \citep{Sklar} states that every 
multivariate distribution $F$ with marginals $F_1$, $F_2$,\ldots,$F_d$ can be written as
\begin{equation}\label{sklar}
F(x_1,\ldots,x_d) = C(F_1(x_1),F_2(x_2),....,F_d(x_d)),
\end{equation}
for some appropriate $d$-dimensional copula $C$. In fact, the copula from \eqref{sklar} has the expression
\begin{equation*}
C(u_1,\ldots,u_d) = F(F_1^{-1} (u_1),F_2^{-1} (u_2),\ldots,F_d^{-1} (u_d)),
\end{equation*}
where the $F_j^{-1}$s are the inverse distribution functions of the marginals. While other copulas may be used, the Gaussian copula has the benefit that we may use the analytical expressions for the conditionals in \eqref{mu} 
and \eqref{cov}.

Assuming a Gaussian copula, we may thus use the following procedure for generating samples from  $p(\boldsymbol{x}_{\bar{\mathcal{S}}} |\boldsymbol{x}_{\mathcal{S}}=\boldsymbol{x}_{\mathcal{S}}^*)$:
\begin{itemize}
	\item Convert each marginal $X_j$ of the feature distribution $\boldsymbol{X}$ to a Gaussian feature $V_j$ by $V_j = \Phi^{-1}(\hat{F}_j(X_j))$, where $\hat{F}_j$ is the empirical distribution function of marginal $j$.
	\item Assume that $\boldsymbol{V}$ is distributed according to a multivariate Gaussian\footnote{The quality of this assumption will depend on how close the Gaussian copula is to the true copula.}, 
	and sample from the conditional distribution $p(\boldsymbol{v}_{\bar{\mathcal{S}}} |\boldsymbol{v}_{\mathcal{S}}=\boldsymbol{v}_{\mathcal{S}}^*)$ using the method described in Section \ref{gaussMet}.
	\item Convert the margins $V_j$ in the conditional distribution to the original distribution using $\hat{X}_j = \hat{F}^{-1}_j(\Phi(V_j))$. 
\end{itemize}
With a series of samples generated as described above, the integral in \eqref{eq:int} is finally approximated by \eqref{expectationIndep}.

\subsection{Empirical conditional distribution}\label{empMet}

If both the dependence structure and the marginal distributions of $\boldsymbol{x}$ are very far from the Gaussian, neither of the two aforementioned methods will work very well. For such situations, we propose a
non-parametric approach. The classical method for non-parametric density estimation is the kernel estimator \citep{rosenblatt1956}, which in the  decades following  its introduction has been refined and developed in
many directions, see for example \cite{Holmes2010, bertin2016, izbicki2017}. However, the kernel estimator suffers greatly from the curse of dimensionality, which quickly inhibits its use in multivariate problems.
Moreover, very few methods exist for the non-parametric estimation of conditional densities, especially when either $\boldsymbol{x}_{\mathcal{S}}$ or $\boldsymbol{x}_{\bar{\mathcal{S}}}$ 
are not one-dimensional. Finally, most kernel estimation approaches gives a non-parametric density estimate, only, while we need to be able to generate samples from the estimated distribution. 

Hence, we have developed an empirical conditional approach, inspired by the Nadaraya-Watson estimator \citep{Bierens94}, to sample approximately from $p(\boldsymbol{x}_{\bar{\mathcal{S}}}|\boldsymbol{x}_{\mathcal{S}}^*)$. The method, which is motivated by the idea that samples 
$(\boldsymbol{x}_{\bar{\mathcal{S}}},\boldsymbol{x}_{\mathcal{S}})$ with $\boldsymbol{x}_{\mathcal{S}}$ close to $\boldsymbol{x}_{\mathcal{S}}^*$ are informative about the conditional distribution $p(\boldsymbol{x}_{\bar{\mathcal{S}}}|\boldsymbol{x}_{\mathcal{S}}^*)$, 
consists of the following steps:

\begin{enumerate}
	\item Compute the distance between the instance $\boldsymbol{x}^*$ to be explained and all training instances $\boldsymbol{x}^i, i=1,\ldots, n_{\text{train}}$. The distance between $\boldsymbol{x}^*$ and instance $i$
	is computed as
	\begin{equation}\label{Maha}
	D_{\mathcal{S}}(\boldsymbol{x}^*,\boldsymbol{x}^i)  = \sqrt{\frac{(\boldsymbol{x}_{\mathcal{S}}^* -\boldsymbol{x}_{\mathcal{S}}^i)^T\Sigma_{\mathcal{S}}^{-1}(\boldsymbol{x}_{\mathcal{S}}^* -\boldsymbol{x}_{\mathcal{S}}^i)}{|\mathcal{S}|}},
	\end{equation}
	where $\Sigma_{\mathcal{S}}$ is the sample covariance matrix for the $n_{\text{train}}$ instances of $\boldsymbol{x}_{\mathcal{S}}$. That is, when we compute the distance we only use the elements in the subset $\mathcal{S}$. \eqref{Maha} may be viewed as 
	a scaled version of the Mahalanobis distance \citep{Mahalanobis36}.
	\item Compute weights for all training instances $\boldsymbol{x}^i, i=1,\ldots, n_{\text{train}}$ from the distances similarly to a Gaussian distribution kernel:
	\[ w_{\mathcal{S}}(\boldsymbol{x}^*,\boldsymbol{x}^i) = \exp\left(-\frac{D_{\mathcal{S}}(\boldsymbol{x}^*,\boldsymbol{x}^i)^2}{2\sigma^2}\right),\]
	where $\sigma$ may be viewed as a smoothing parameter or bandwidth that needs to be specified.
	\item Sort the weights $w_{\mathcal{S}}(\boldsymbol{x}^*,\boldsymbol{x}^i)$ in increasing order, and let $\boldsymbol{x}^{[k]}$ be the training instance corresponding to the $k$th largest weight. 
	\item Approximate the integral in \eqref{eq:int} with a weighted version of \eqref{expectationIndep}: 
	\[v_{\text{condKerSHAP}}(\mathcal{S}) = \frac{\sum_{k=1}^K w_{\mathcal{S}}(\boldsymbol{x}^*,\boldsymbol{x}^{[k]}) f(\boldsymbol{x}_{\bar{\mathcal{S}}}^{[k]},\boldsymbol{x}_{\mathcal{S}}^*)}{\sum_{k=1}^K w_{\mathcal{S}}(\boldsymbol{x}^*,\boldsymbol{x}^{[k]})}.\]
\end{enumerate}
Note that we could have used \eqref{expectationIndep}, with the $\boldsymbol{x}_{\bar{\mathcal{S}}}^{k}$ sampled (with replacement) from the training data with weights $w_{\mathcal{S}}(\boldsymbol{x}^i), i=1,\ldots,n_{\text{train}}$. Our approach is, however, 
more sampling effective as it uses each training observation only once, and uses their weights in the integral computation, rather than as input for the sampling only.

The number of samples $K$ to be used in the approximate prediction in step 4 can for instance be chosen such that most of the total sum of weights is accounted for by the $K$ largest weights:
\begin{align}
K = \min_{L \in \mathbb{N}}\left\{ \frac{\sum_{k=1}^L w_{\mathcal{S}}(\boldsymbol{x}^*,\boldsymbol{x}^{[k]})}{\sum_{i=1}^{n_{\text{train}}} w_{\mathcal{S}}(\boldsymbol{x}^*,\boldsymbol{x}^{i})} > \eta\right\}, \label{eq:KKK}
\end{align}
where $\eta$ is set to for instance 0.9. If $K$ in \eqref{eq:KKK} exceeds a certain limit, for instance 5,000, it might be set to that limit. 

Essentially all kernel based estimation procedures (such as kernel density estimation) requires selection of one or more bandwidth parameters. Our method is no exception. 
The choice of the bandwidth parameter $\sigma$ may be viewed as a bias-variance trade-off. A small $\sigma$ puts most of the weight to a few of the closest training observations and thereby gives low bias, but high variance. A large $\sigma$ spreads 
the weight to a higher number of (more distant) training observations and thereby gives high bias, but low variance. Typically, when the features are highly dependent, a small $\sigma$ is needed such that the bias does not dominate. When the features 
are essentially independent, there is no bias and a larger $\sigma$ is preferable. As $\sigma \rightarrow \infty$ our method approximates the original Kernel SHAP method in Section \ref{expectation1}.


By viewing the estimation of $E[f(\boldsymbol{x})|\boldsymbol{x}_{\mathcal{S}}=\boldsymbol{x}_{\mathcal{S}}^*]$ as a regression problem with response $f(\boldsymbol{x}_{\bar{\mathcal{S}}}^{i},\boldsymbol{x}_{\mathcal{S}}^*)$ and 
covariates $\boldsymbol{x}_{\mathcal{S}}^{i}, i=1,\ldots,n_{\text{train}}$, it turns out that our empirical conditional distribution approach (with $K=n_{\text{train}}$) is equivalent to the Nadaraya-Watson estimator \citep{Bierens94}.
\cite{Hurvich98} have developed a small-sample-size corrected version of Akaike information criterion (AICc) to select bandwidth parameters in such nonparametric regression problems. The connection to the Nadaraya-Watson 
estimator allows us to apply the AICc directly to select $\sigma$.

The strategy used in AICc to find a suitable smoothing parameter is to choose the $\sigma$ 
which is the minimizer of 
\begin{align}
\text{AICc} = \log(\hat{\tau}^2) + \Phi(H), \label{eq:AICc}
\end{align}
where 
\[\hat{\tau}^2=\frac{1}{n_{\text{train}}} \sum_{i=1}^{n_{\text{train}}} \left(f(\boldsymbol{x}_{\bar{\mathcal{S}}}^{i},\boldsymbol{x}_{\mathcal{S}}^*)-\frac{\sum_{j=1}^{n_{\text{train}}} w_{\mathcal{S}}(\boldsymbol{x}^j,\boldsymbol{x}^i)f(\boldsymbol{x}_{\bar{\mathcal{S}}}^{j},\boldsymbol{x}_{\mathcal{S}}^*)}       
{\sum_{j=1}^{n_{\text{train}}} w_{\mathcal{S}}(\boldsymbol{x}^j,\boldsymbol{x}^i)}\right)^2, \]
and 
\[\Phi(H) = \frac{1+\text{tr}(H)/n_{\text{train}}}{1-(\text{tr}(H)+2)/2}.\]
Here $H$ is the $n_{\text{train}} \times n_{\text{train}}$ matrix with indexes 
\[H_{i,j} = \frac{w_{\mathcal{S}}(\boldsymbol{x}^j,\boldsymbol{x}^i)}{\sum_{l=1}^{n_{\text{train}}} w_{\mathcal{S}}(\boldsymbol{x}^l,\boldsymbol{x}^i)}\] 
commonly called the smoother or hat matrix.

Thus, to select $\sigma$ we compute the $\text{AICc}$ in \eqref{eq:AICc} for various $\sigma$ values and select the $\sigma$ corresponding to the smallest AICc value. This should be done for all subsets $\mathcal{S}$ and every new observation $\boldsymbol{x}^*$ to be explained,
meaning that it is a computationally intensive approach. To reduce the computational burden, we have experimented with different approximations, and ended up with one where $\sigma$ is assumed to have the same value for all subsets $\mathcal{S}$ of the same size $|\mathcal{S}|$. 
In this approach, which is denoted the approximate AICc method in Section \ref{experiment}, the sum of the AICc values for all subsets of the same size is minimized instead of the AICc values for each subset.

Even the approximate AICc method is quite time consuming. Hence, in Section \ref{experiment} we have also experimented with a fixed $\sigma=0.1$ for all subsets $\mathcal{S}$.


\subsection{A combined approach}\label{combapp}

When performing the experiments to be described in Section \ref{experiment}, it turned out that the empirical conditional distribution method works very well if the dimension of $\boldsymbol{x}_{\mathcal{S}} \leq D^*$, where $D^*$ is 
a small number, while it is outperformed by the multivariate Gaussian method and the Gaussian copula method if we condition on more features. This is in accordance with previous literature, see for instance \cite{Izbicki17} and references therein. 
Very few papers
attempt to estimate $f(\boldsymbol{z}|\boldsymbol{x})$ when $\boldsymbol{x}$ has more than $D=3$ dimensions, even if $\boldsymbol{z}$ is one-dimensional. In higher dimensions, the previously proposed methods typically 
rely on a prior dimension reduction step, which can result in significant loss of information.
Hence, it might be wise to combine the empirical approach with either the multivariate Gaussian or the Gaussian copula approach, simulating the conditional distributions for which $\mbox{dim}(\boldsymbol{x}_{\mathcal{S}}) \leq D^*$ using the empirical 
method, and all other conditional distributions using the parametric method. In this combined approach we at least partly avoid the curse of dimensionality, since the empirical approach is used only when conditioning on a few features and
the conditional distributions can be analytically computed for the Gaussian approach.  To determine $D^*$, one may use for instance cross validation. In our experiments we have, however, determined $D^*$ using an ad-hoc procedure.

\section{Experiments} \label{experiment}

A problem with evaluating prediction explanation methods is that there generally is no ground truth. Hence, to verify that our approaches are more accurate than the original Kernel SHAP method described in Section \ref{expectation1}
when we have dependent features, we have to turn to simulated data for which we may compute the true Shapley values.

As computing the exact Shapley values for a single prediction requires solving $O(2^M)$ integrals of the type in \eqref{eq:int}, which are of dimension $1$ to $M-1$, we cannot perform accurate experiments in high dimensional settings. We will however perform experiments in a low dimensional ($M=3$) 
and moderate dimensional ($M=10$) setting, using various multivariate distributions for the features $\boldsymbol{x}$, sampling models for $y|\boldsymbol{x}$, and forms of the predictive model $f()$. The experiments with $M=3$ and $M=10$ are treated in Sections \ref{dim3} 
and \ref{dim10}, respectively. Due to the low/moderate dimension of the features in these experiments, it is computationally tractable to use the exact version of Kernel SHAP in \eqref{eq:KS.formula}.  Hence, we do not have to turn to the approximation in  \eqref{eq:KS.formula.approx}.

As the AICc dependent approximation methods are directly dependent on the sampled training set, we run the experiments in 10 batches.  In each batch, we sample a new training set of size $n_{\text{train}}=2,000$, and use the model fitted to those training data to explain predictions in 
a test set of size 100. These means that the quality of the conditional expectation approximations is measured based on a total of $n_{\text{test}}=10 \cdot 100 = 1,000$ test observations. Sampling new training data for each batch also reduces the influence of the exact form of the 
fitted predictive model, compared to using a single training set across all simulations.


The Shapley value approximations we compare with the original Kernel SHAP method (original) are:
\begin{itemize}
	\item The Kernel SHAP with the Gaussian conditional distribution (Gaussian)
	\item The Kernel SHAP with the Gaussian copula and empirical margins (copula)
	\item The Kernel SHAP with the empirical conditional distribution determining $\sigma$ with exact AICc (empirical-AICc-exact) 
	\item The Kernel SHAP with the empirical conditional distribution determining $\sigma$ with approximate AICc (empirical-AICc-approx) 
	\item The Kernel SHAP with the empirical conditional distribution setting $\sigma=0.1$ for all conditional distributions (empirical-0.1) 
	\item The Kernel SHAP with the combined approach using the empirical approach for subsets with dimension $\leq 3$ and the Gaussian approach otherwise (empirical-0.1+Gaussian and empirical-AICc-approx+Gaussian)
	\item The Kernel SHAP with the combined approach using the empirical approach for subsets with dimension $\leq 3$ and the copula approach otherwise (empirical-0.1+copula and empirical-AICc-approx+copula)
\end{itemize}
Due to computational complexity, the empirical-AICc-exact method is only used in the 3 dimensional experiments. Furthermore, the combined approaches are only used in the 10 dimensional experiments.
For the experiments where XGBoost is used to fit the predictive model, we will also include the so-called TreeSHAP method \citep{Lundberg2} in the comparison.

\subsection{Evaluation measures}

To quantify the accuracy of the different methods, we rely on the mean absolute error (MAE) of the Shapley value approximations, averaged over all features and all test samples, that is
\begin{equation}
\text{MAE}(\text{method }q) = \frac{1}{Mn_{\text{test}}}\sum_{j=1}^{M}\sum_{i=1}^{n_{\text{test}}} |\phi^{(i)}_{j,\text{true}} - \phi^{(i)}_{j,q}|, \notag
\end{equation}
where $\phi^{(i)}_{j,q}$ denotes the Shapley value of feature $j$, for prediction $i$, and computed with approximation method $q$, while $\phi^{(i)}_{j,\text{true}}$ is the corresponding true value. 
In order to determine the superiority of the various proposed methods over the original Kernel SHAP method, 
we rely on the so-called skill score \citep{gneiting2007strictly} associated with the aforementioned MAE. The skill score for method $q$ takes the form 
\begin{align}
\text{Skill}(\text{MAE, method } q) = \frac{\text{MAE}(\text{method }q) - \text{MAE}(\text{original})}{\text{MAE}(\text{optimal}) - \text{MAE}(\text{original})} = 
1- \frac{\text{MAE}(\text{method }q)}{\text{MAE}(\text{original})}, \notag
\end{align}
where $\text{MAE}(\text{optimal})$ is the MAE of an optimal method, being equal to zero. The skill score measures the superiority of a method compared to the reference method (here the original Kernel SHAP). It is 
standardized in such a way that it takes the value 1 for a perfect approximation and 0 for the reference method. When the approximation method is worse than the reference method, the skill score becomes negative.

To compare the methods on equal terms, all methods are restricted to use $K=1000$ samples from the training set for each feature combination and test observation.

\subsection{Dimension 3} \label{dim3}

For the three dimensional setting, we will use two different sampling models for $y|\boldsymbol{x}$ (linear and piecewise constant), and combine these with three different multivariate distributions for the features $\boldsymbol{x}$ (Gaussian, Generalized Hyperbolic distribution and Gaussian mixture), such that we get a total of six experimental setups A-F. In Sections
\ref{sec:feature.dist.1}---\ref{sec:feature.dist.3} we describe the multivariate feature distributions, while the sampling models are discussed in Sections \ref{sec:samp.models.1} and \ref{sec:samp.models.2}. Finally, Section \ref{sec:results} contains the
results. 

The noise term $\varepsilon_i$ is common for all experients and assumed to follow the distribution $\varepsilon_i \stackrel{\text{d.}}{=} \varepsilon \sim \text{N}(0,0.1^2)$.
Due to the low dimension, the exact Shapley value in \eqref{eq:int} can be computed using numerical integration. 

\subsubsection{Gaussian distributed features} \label{sec:feature.dist.1}

The first feature distribution $p(\boldsymbol{x})$ we shall consider is
a multivariate Gaussian distribution $p(\boldsymbol{x}) = \text{N}_3(\boldsymbol{0},\boldsymbol{\Sigma}(\rho))$, where the covariance matrix $\boldsymbol{\Sigma}(\rho)$ takes the form
\begin{align}
\boldsymbol{\Sigma}(\rho) = \begin{bmatrix} 
1 & \rho & \rho\\
\rho & 1 & \rho\\
\rho & \rho & 1
\end{bmatrix}. \label{eq:covarmat}
\end{align}
In the various experiments, the correlation coefficient $\rho$ varies between 0 and 0.98, representing an increasing positive correlation among the features.

\subsubsection{Skewed and heavy-tailed  distributed features} \label{sec:feature.dist.2}

The second feature distribution $p(\boldsymbol{x})$ to be considered is the Generalized Hyperbolic(GH)-distribution. 
Following \cite{Browne}, a random vector $\boldsymbol{X}$ is said to follow a GH-distribution with index parameter $\lambda$, concentration parameter $\omega$, location vector $\boldsymbol{\mu}$, dispersion matrix  $\boldsymbol{\Sigma}$, and skewness vector $\boldsymbol{\beta}$, denoted by 
$\boldsymbol{X}\sim \mbox{GH}(\lambda,\omega,\boldsymbol{\mu},\boldsymbol{\Sigma},\boldsymbol{\beta})$, if it can be represented by
\[ \boldsymbol{X} = \boldsymbol{\mu} + W\,\boldsymbol{\beta} + \sqrt{W}\,\boldsymbol{U},\]
where $W\sim \mbox{GIG}(\lambda,\omega,\omega)$, $\boldsymbol{U}\sim N(\boldsymbol{0},\boldsymbol{\Sigma})$ and $W$ is independent of $\boldsymbol{U}$. GIG is the Generalised Inverse Gaussian distribution introduced by \cite{Good}. 
Appendix \ref{app1} contains more details on this distribution. We use the following parameter values in our experiments:
\begin{align}
\lambda &= 1 \notag \\
\omega &= 0.5 \notag \\
\boldsymbol{\Sigma} &= \boldsymbol{\Sigma}(0) = \text{diag}(\boldsymbol{1}) \notag \\
\boldsymbol{\beta} &= (1/4)\kappa*\boldsymbol{1} \notag \\
\boldsymbol{\mu} &= \boldsymbol{0} - \text{E}[W](1/4)\kappa, \notag
\end{align}
where the skewness coefficient $\kappa$ varies from 1 to 10 in different experiments, resulting in an increasingly more skewed, heavy-tailed and positively correlated distribution.
$\text{E}[W]$ denotes the mean of the GIG distribution, equal to approximately 4.56 for the above parameter values. The special form of $\boldsymbol{\mu}$ is chosen such that the GH distribution has mean zero.

\subsubsection{Multi-modal distributed features} \label{sec:feature.dist.3}
The last feature distribution is a mixture of two Gaussian distribution with different means. That is, 
\[ p(\boldsymbol{x}) = \sum_{k=1}^2 \pi_k \text{N}(\boldsymbol{\mu}_k(\gamma), \boldsymbol{\Sigma}(0.2)),\]
with mixture probabilities $\pi_1=\pi_2=0.5$ and mean functions $\boldsymbol{\mu}_1(\lambda) = -\boldsymbol{\mu}_2(\lambda) = \\ \gamma \cdot (1,-0.5,1)^\top$. The covariance matrix $\boldsymbol{\Sigma}$ is assumed to be on the form in \eqref{eq:covarmat}. 
The $\gamma$ parameter represents the distance between the two Gaussian distributions, and will range from 0.5 to 10 in the different experiments.

\subsubsection{Linear model} \label{sec:samp.models.1}

The first sampling model $y|\boldsymbol{x}$ is a simple linear model:
\begin{align}
y = g(\boldsymbol{x}) = x_1 + x_2 + x_3 + \varepsilon. \notag
\end{align}
Data from such a distribution will be modelled by fitting the parameters $\beta_j,j=0,\ldots,3$ in the linear model
\begin{equation}
f(\boldsymbol{x}) = \beta_0 + \beta_1\,x_1 + \beta_2\,x_2 + \beta_3\,x_3 + \varepsilon,  \notag
\end{equation}
using ordinary linear regression.

\subsubsection{Piecewise constant model} \label{sec:samp.models.2}

\begin{figure}[ht]
	\begin{center}
		\includegraphics[width=0.7\linewidth]{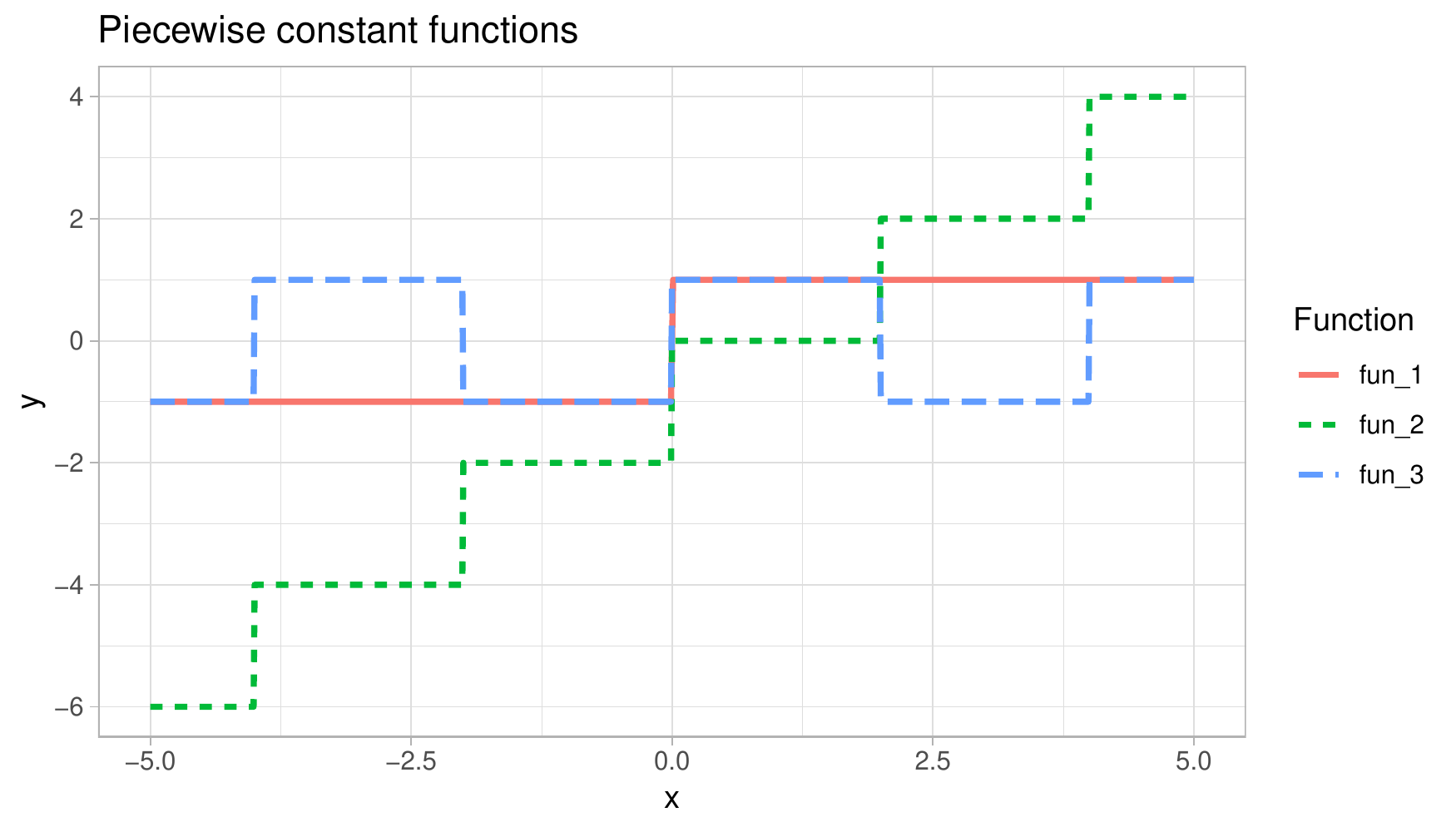}
	\end{center}
	\caption{Piecewise constant functions used in experiment D, E and F.
		\label{fig:piecewise_constant_funcs}}
\end{figure}

The second sampling model $y|\boldsymbol{x}$ is a piecewise constant model constructed by summing three different piecewise constant functions:
\begin{align}
y = g(\boldsymbol{x}) = \text{fun}_1(x_1) + \text{fun}_2(x_2) + \text{fun}_3(x_3) + \varepsilon. \notag
\end{align}
The functions $\text{fun}_1, \text{fun}_2$, and $\text{fun}_3$ 
are displayed in Figure \ref{fig:piecewise_constant_funcs}.
We fit the model $g(\boldsymbol{x})$ with the XGBoost framework \citep{Chen2016} using default values of all hyperparameters, the histogram tree learning method (\verb,tree_method="hist",) and 50 boosting iterations.

\subsubsection{Results}\label{sec:results}

\begin{figure}[ht]
	\begin{center}
		\includegraphics[width=1.0\linewidth]{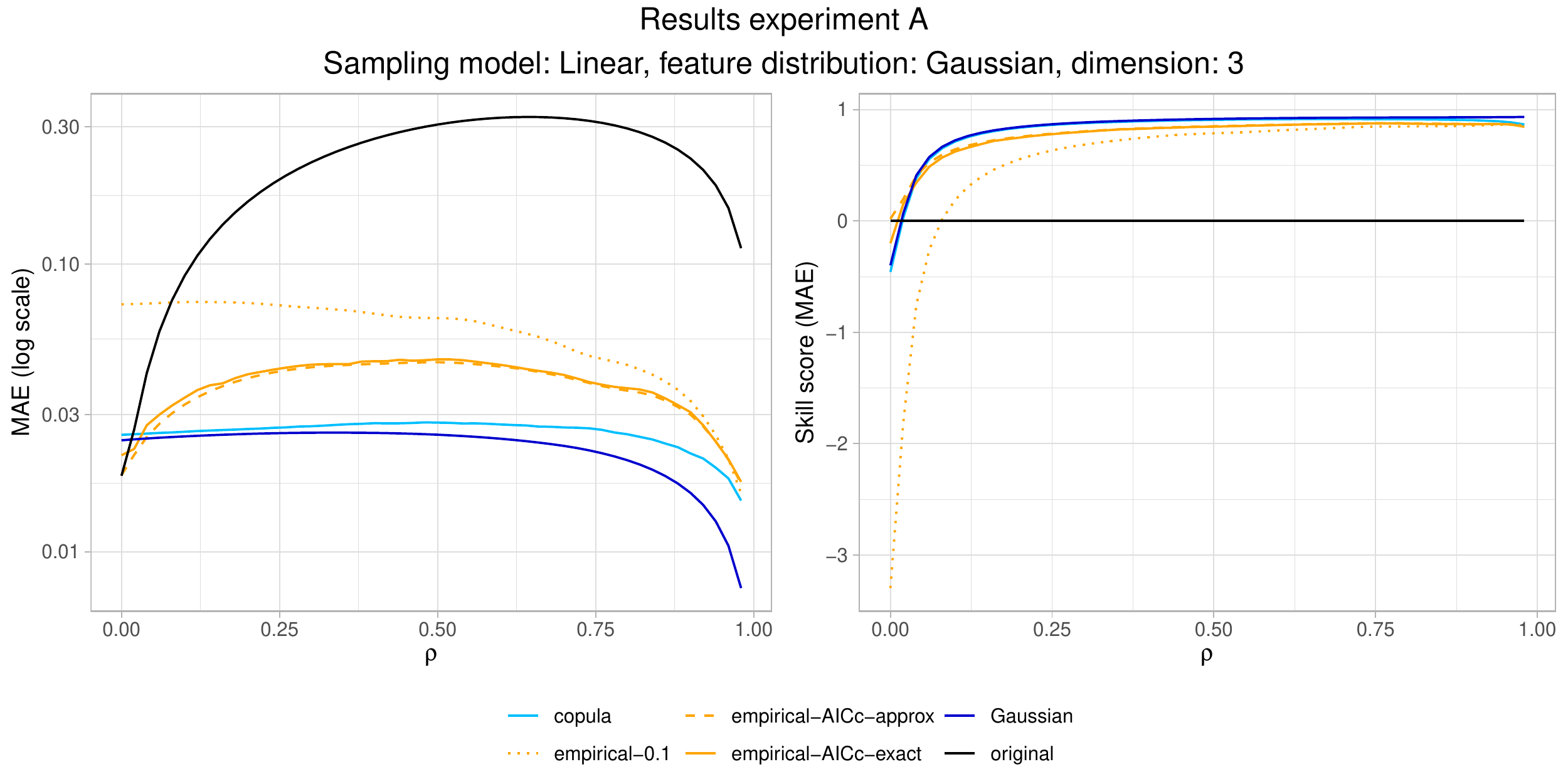}
	\end{center}
	\caption{Dimension 3: MAE and skill score for linear model with Gaussian features.
		\label{fig:exA3}}
\end{figure}

\begin{figure}[ht]
	\begin{center}
		\includegraphics[width=1.0\linewidth]{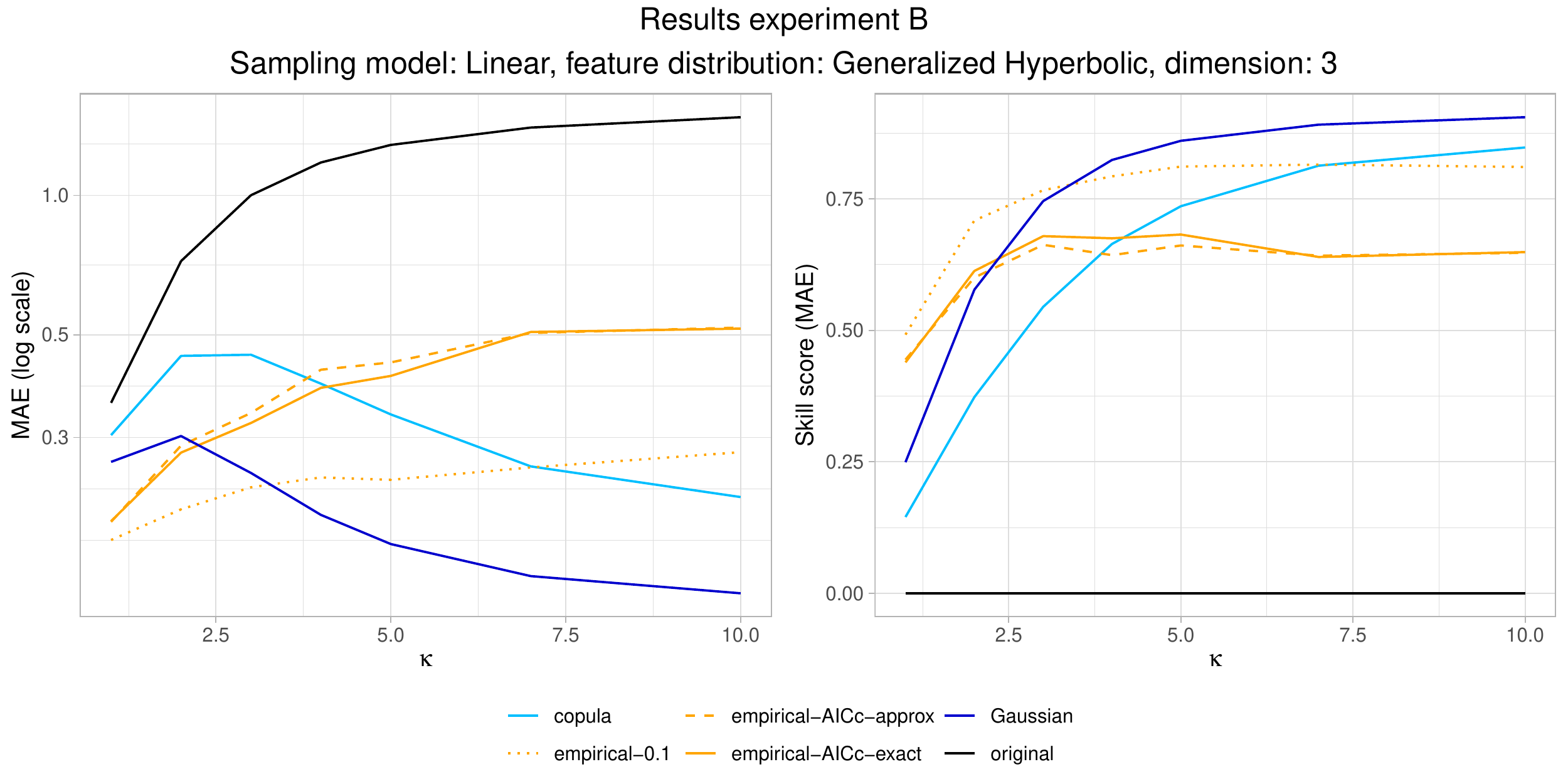}
	\end{center}
	\caption{Dimension 3: MAE and skill score for linear model with GH-distributed features.
		\label{fig:exB3}}
\end{figure}

\begin{figure}[ht]
	\begin{center}
		\includegraphics[width=1.0\linewidth]{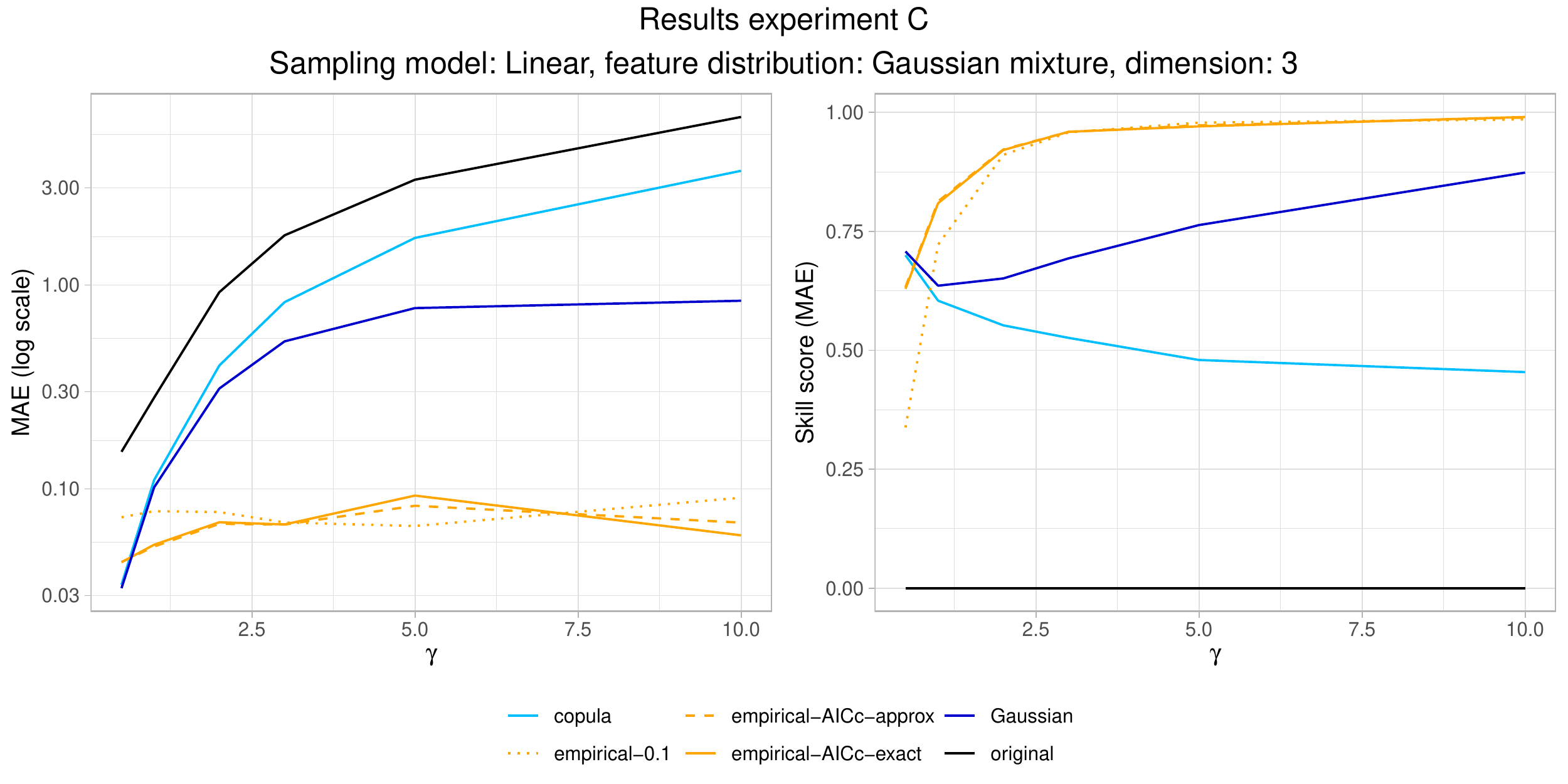}
	\end{center}
	\caption{Dimension 3: MAE and skill score for linear model with Gaussian mixture distributed features.
		\label{fig:exC3}}
\end{figure}

\begin{figure}[ht]
	\begin{center}
		\includegraphics[width=1.0\linewidth]{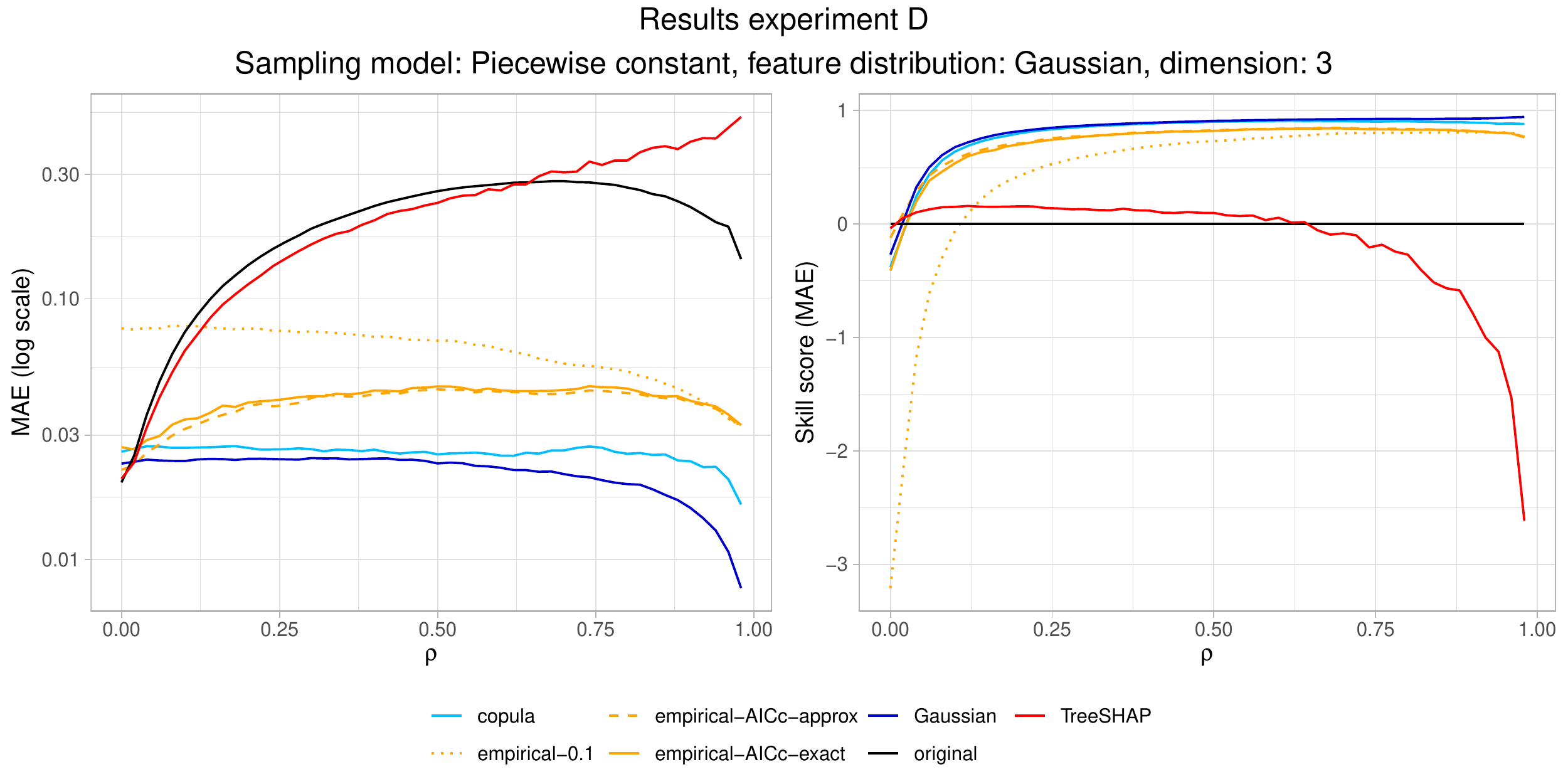}
	\end{center}
	\caption{Dimension 3: MAE and skill score for piecewise constant model with Gaussian features.
		\label{fig:exD3}}
\end{figure}

\begin{figure}[ht]
	\begin{center}
		\includegraphics[width=1.0\linewidth]{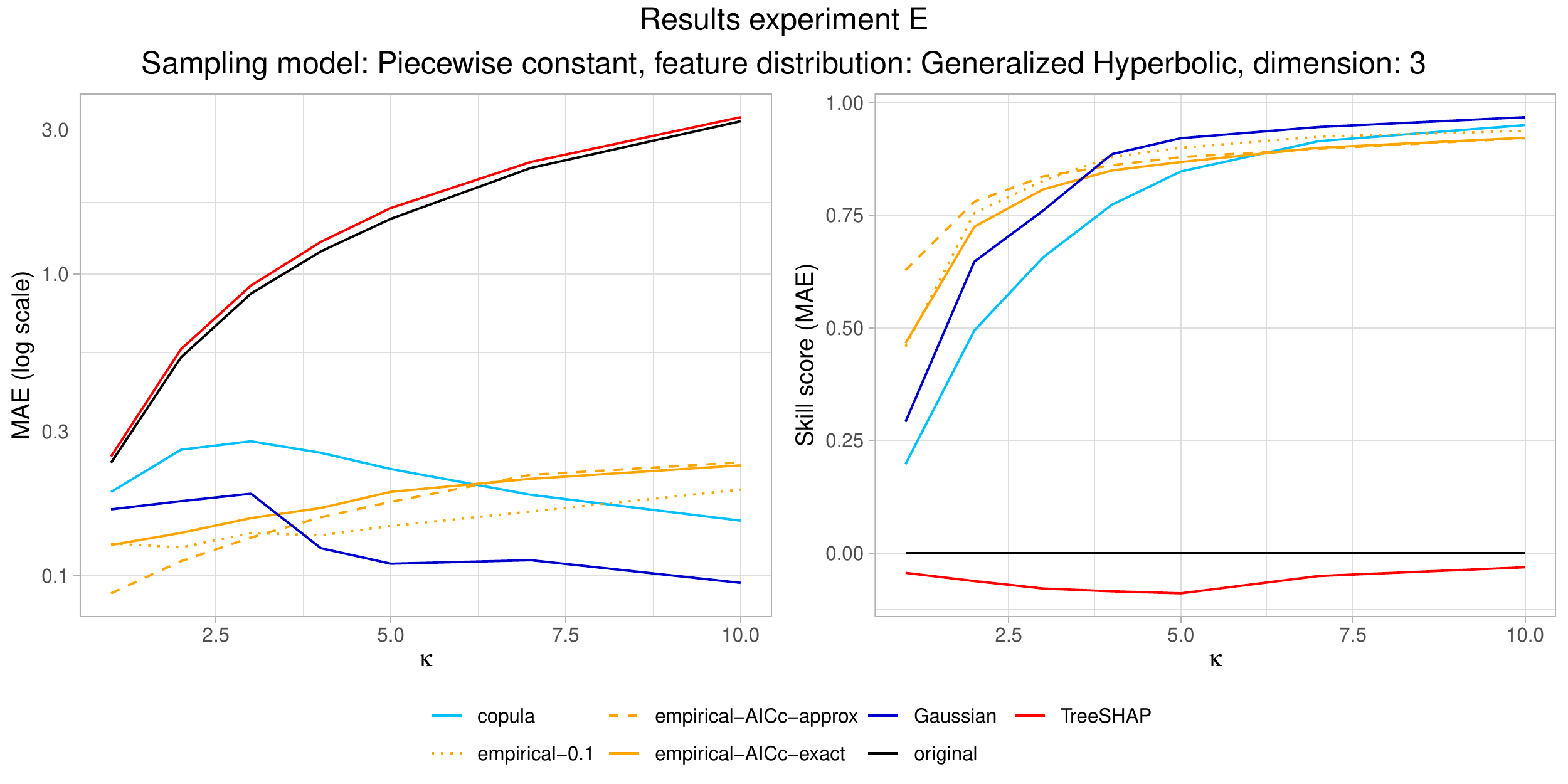}
	\end{center}
	\caption{Dimension 3: MAE and skill score for piecewise constant model with GH-distributed features.
		\label{fig:exE3}}
\end{figure}

\begin{figure}[ht]
	\begin{center}
		\includegraphics[width=1.0\linewidth]{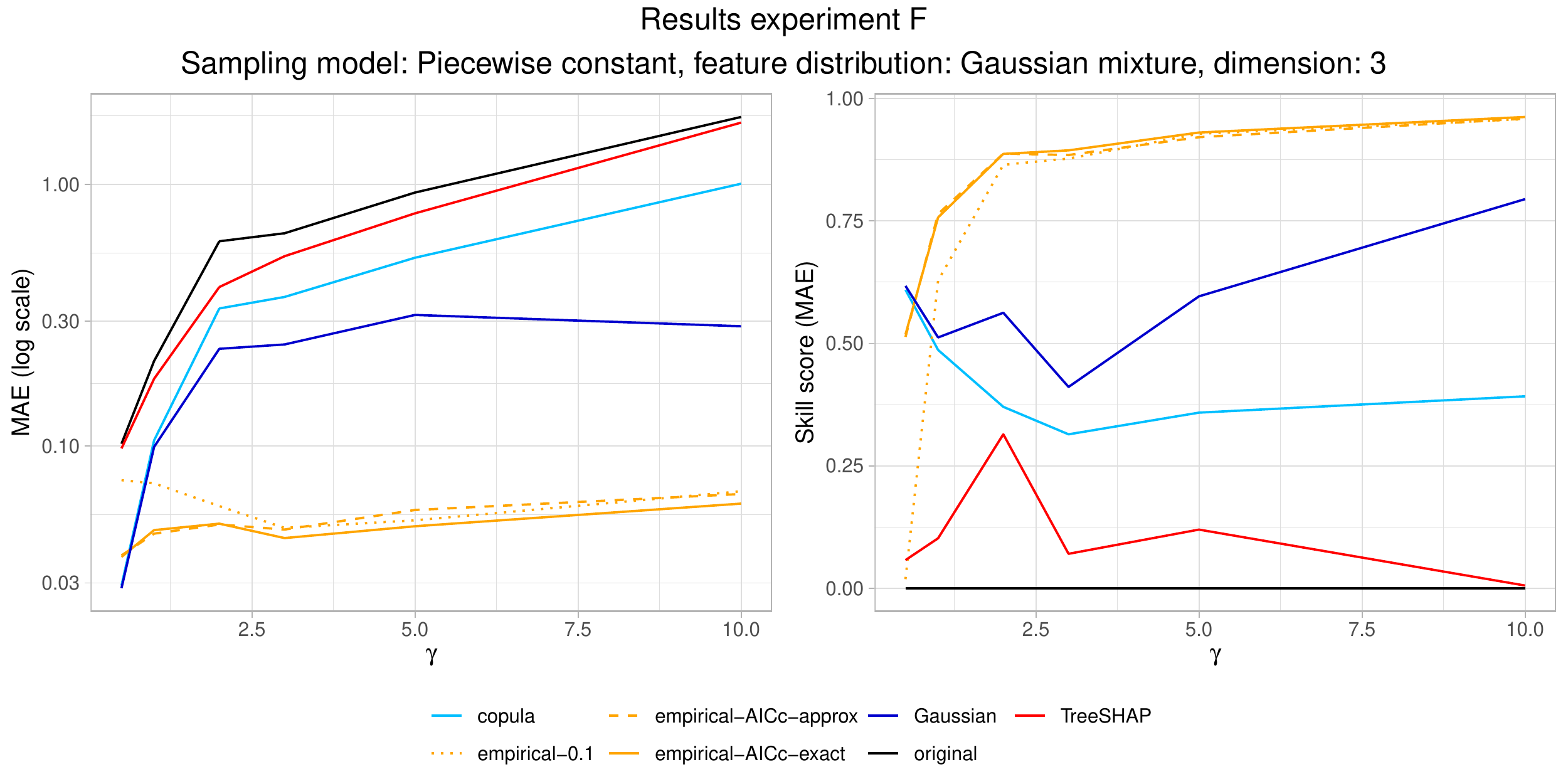}
	\end{center}
	\caption{Dimension 3: MAE and skill score for piecewise constant model with Gaussian mixture distributed features.
		\label{fig:exF3}}
\end{figure}

The results from the 3D-experiments are visualized in Figures \ref{fig:exA3}---\ref{fig:exF3}. In experiment A we have used a linear sampling model and Gaussian features. As seen from Figure  \ref{fig:exA3}, the original Kernel 
SHAP method works well when the features are independent, but it is outperformed by all other
methods when $\rho$ is greater than 0.05. The Gaussian model generally shows the best performance. It should also be noted that the AICc methods for determining $\sigma$ works better than using the fixed value of 0.1.

In experiment B, we still have the linear sampling model, but now we have skewed and heavy-tailed features. Like for the previous experiment, the original Kernel SHAP method is outperformed by all other approaches. 
As shown in Figure \ref{fig:exB3}, when $\kappa$ increases, the MAE of the copula and Gaussian methods is reduced. This might be due to the increased variance of the features that comes as a by-product of increasing the $\kappa$ parameter. 
For this experiment, the empirical approach with a fixed $\sigma=0.1$ performs uniformly better than those based on AICc.

In experiment C, we have the same sampling model, but now with bimodal Gaussian mixture distributed features. Again, all our methods perform uniformly better than the original Kernel SHAP method. Figure \ref{fig:exC3} further
shows that the empirical methods outperform the Gaussian and copula methods, especially when $\gamma$ (that is, the distance between the modes of the feature distribution) is large.

In experiments D-F we use the same feature distributions as in A-C, but for these experiments we use a piecewise constant model instead of the linear. The results are largely the same as in the linear model case.
The original Kernel SHAP method is outperformed by all other approaches. Further the Gaussian model is best when we have Gaussian features, while the empirical approaches are best when the feature distribution is bi-modal. 
In the case with skewed and heavy-tailed features, the Gaussian and copula methods again seem to be preferable for larger values of $\kappa$.  

For the experiments with the piecewise constant model, we have used the TreeSHAP method in addition to the other ones. As shown in Figures \ref{fig:exD3}-\ref{fig:exF3}, the performance of this method is just
slightly better than that of the original kernel SHAP method for experiments D and E, and worse in experiment F. This is surprising, since the TreeSHAP method is supposed to handle dependence better than the original kernel SHAP method.

Overall, our 3 dimensional experiments clearly shows that it is important to account for the dependence between the features when computing the Shapley values. Which of the suggested methods that is best, depends on the underlying feature distribution, 
Further, since the results for our empirical method using the approximate AICc version are fairly similar (and in some cases even better) to those using the slower exact version, we recommend the former.

\clearpage

\subsection{Dimension 10}\label{dim10}

In the 10 dimensional case we have restricted ourselves to three types of experiments. In the first two, we use Gaussian distributed features on the same form as described for the 3 dimensional case in Section \ref{sec:feature.dist.1}. 
That is, $p(\boldsymbol{x}) = \text{N}_{10}(\boldsymbol{0},\boldsymbol{\Sigma}(\rho))$, where $\boldsymbol{\Sigma}(\rho)$ is the 10 dimensional extension of \eqref{eq:covarmat} such that all features have variance 1 and 
the pairwise correlation between any two features is $\rho$.

The first experiment uses a sampling model $y|\boldsymbol{x}$ directly extending the 3 dimensional linear model in Section \ref{sec:samp.models.1}, that is, 
\[ y = g(\boldsymbol{x}) = \sum_{j=1}^9 x_j + \varepsilon,\]
where we use the same error term $\varepsilon_i \stackrel{\text{d.}}{=} \varepsilon \sim \text{N}(0,0.1^2)$ as in the 3 dimensional experiments.
Analogous to the 3 dimensional case, $y$ is modelled by ordinary linear regression 
\[ f(\boldsymbol{x}) = \sum_{j=1}^{10} \beta_jx_j + \varepsilon.\]
Note that even though we have 10 features, one of them has no influence on $y$.

The second experiment extends the 3 dimensional experiment described in Section \ref{sec:samp.models.2}, taking the form
\[ y = g(\boldsymbol{x}) = \sum_{j \,\in \{1,2,3\}} \text{fun}_1(x_j) + 
\sum_{j \,\in \,\{4,5,6\}} \text{fun}_2(x_j) + 
\sum_{j \,\in \,\{7,8,9\}} \text{fun}_3(x_j) + \varepsilon,\]
again excluding the effect of the 10th feature. 
As for the 3 dimensional case, the model $g(\boldsymbol{x})$ is fitted using the XGBoost framework, using the same settings.

In the last experiment, we simulate data from a 10 dimensional GH-distribution with the following parameter values
\begin{align}
\lambda &= 1 \notag \\
\omega &= 0.5 \notag \\
\boldsymbol{\mu} &= (3,3,3,3,3,3,3,3,3,3)\notag \\
\boldsymbol{\Sigma} &= \text{diag}(1,2,3,1,2,3,1,2,3,3)\notag \\
\boldsymbol{\beta} &= (1,1,1,1,1,0.5,0.5,0.5,0.5,0.5). \notag 
\end{align}
The parameter values are chosen as to resemble the feature distribution in our real data example to be described in Section \ref{sec:real}. As sampling model, we use the piecewise constant model described above.  

We didn't use the empirical-AICc-exact approach for any of the 10 dimensional experiments. As previously stated, this approach is computationally intensive. Moreover, the 3 dimensional experiments showed that
the performance of the empirical-AICc-exact and the empirical-AICc-approx approaches were very similar.  For the 3 dimensional experiment with the Generalized Hyperbolic features and piecewise constant model, 
the MAE and skill scores for the three empirical approches were almost equal, meaning that it is not necessary to use the significantly more computational heavy AICc approach. Hence, the only empirical approach used
in the last experiment was the empirical-0.1 method. In the combined approaches, we use the empirical approach for subsets with dimension $\leq 3$, with the bandwidth parameter $\sigma$ either determined by 
the approximate AICc method (in the linear case, only) or fixed to 0.1.

\begin{figure}[ht]
	\begin{center}
		\includegraphics[width=1.0\linewidth]{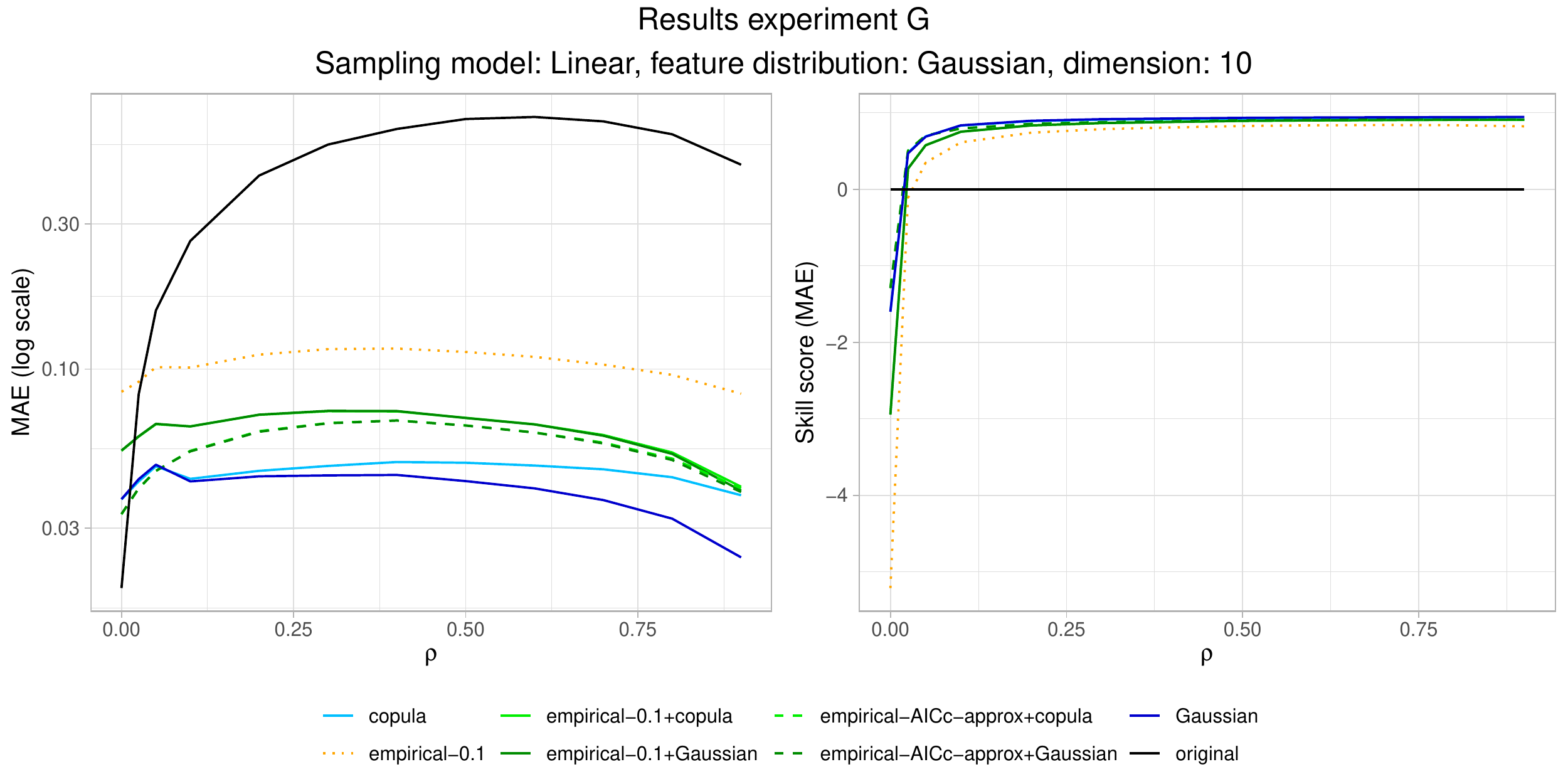}
	\end{center}
	\caption{Dimension 10: MAE and skill score for linear model with Gaussian distributed features.
		\label{fig:exG10}}
\end{figure}

\begin{figure}[ht]
	\begin{center}
		\includegraphics[width=1.0\linewidth]{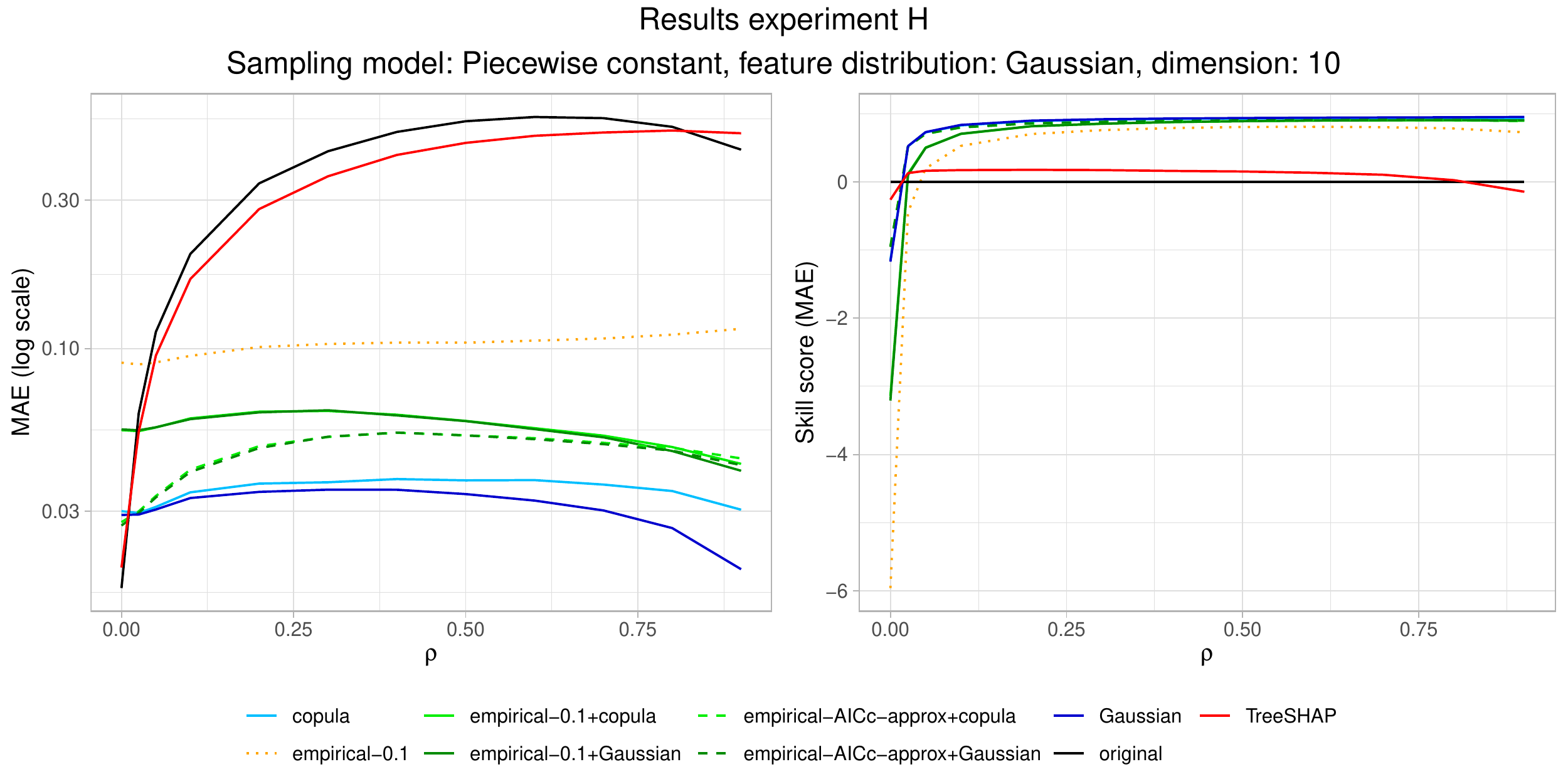}
	\end{center}
	\caption{Dimension 10: MAE and skill score for piecewise constant model with Gaussian distributed features.
		\label{fig:exH10}}
\end{figure}

The results from the simulation experiments are shown in Figures \ref{fig:exG10} and \ref{fig:exH10} and Table \ref{gh10DTab}. While numerical integration was used to compute the exact Shapley values in the 3 dimensional experiments, 
we have to turn to Monte Carlo integration for the 10 dimensional case. From the figures, we see that the results with Gaussian distributed features in dimension 10 are mostly 
the same as for the 3 dimensional counterparts in Figures \ref{fig:exA3} and  \ref{fig:exB3}, with the Gaussian method generally showing the best performance. The combined empirical and Gaussian/copula approaches also works well. 
For the piecewise constant model, the TreeSHAP method behaves as in the 3 dimensional case: slightly better than the original kernel SHAP method for small and medium sized dependence, but worse when 
there is high dependence between the features.

For the skewed and heavy-tailed data, Table \ref{gh10DTab} shows that the empirical-0.1+Gaussian method gives the best performance, having slightly better MAE values and skill scores than the empirical-0.1+copula method.
Finally, like for the other experiments, all our suggested approaches outperform the original Kernel SHAP and the TreeSHAP method. 


\begin{table}
	\begin{center}
		\begin{tabular}{lrr}
			\hline
			Approach & MAE & Skill score \\
			\hline
			original                           & 1.182 & 0.000\\
			Gaussian                           & 0.377 & 0.633\\
			copula                             & 0.526 & 0.504\\
			empirical-0.1                      & 0.307 & 0.737\\
			empirical-0.1+Gaussian             & 0.199 & 0.821\\
			empirical-0.1+Copula               & 0.236 & 0.791\\
			TreeSHAP                           & 1.181 & 0.014\\
			\hline 
		\end{tabular}
		\caption{Dimension 10: MAE and skill score for piecewise constant model with GH-distributed features
			\label{gh10DTab}}
	\end{center}
\end{table}

\subsection{Real data example\label{sec:real}}

\begin{figure}[ht]
	\begin{center}
		\includegraphics[width=0.7\linewidth]{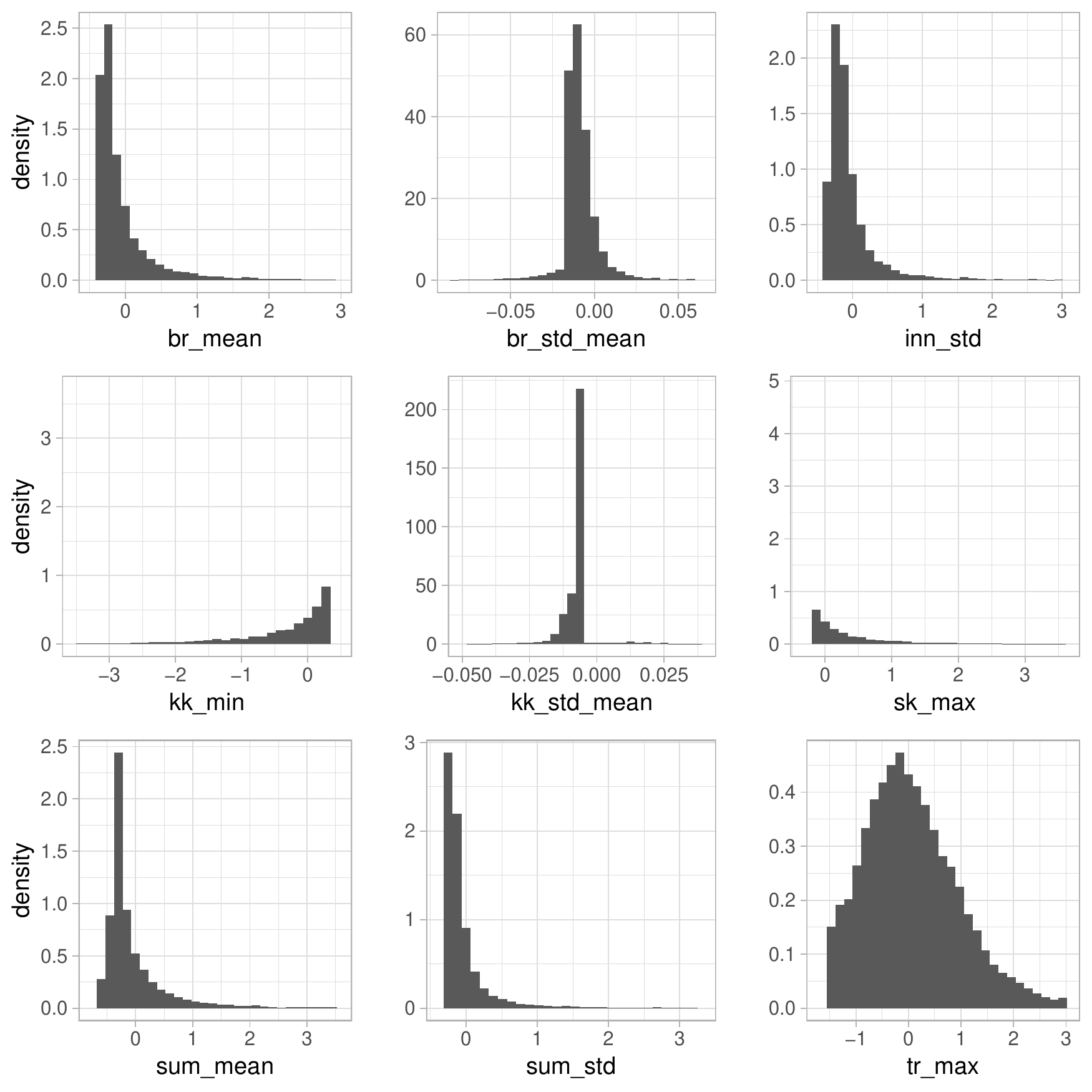}
	\end{center}
	\caption{Histograms for nine of the variables in the real data set. 
		\label{fig:histReal}}
\end{figure}

\begin{figure}[ht]
	\begin{center}
		\includegraphics[width=1.0\linewidth]{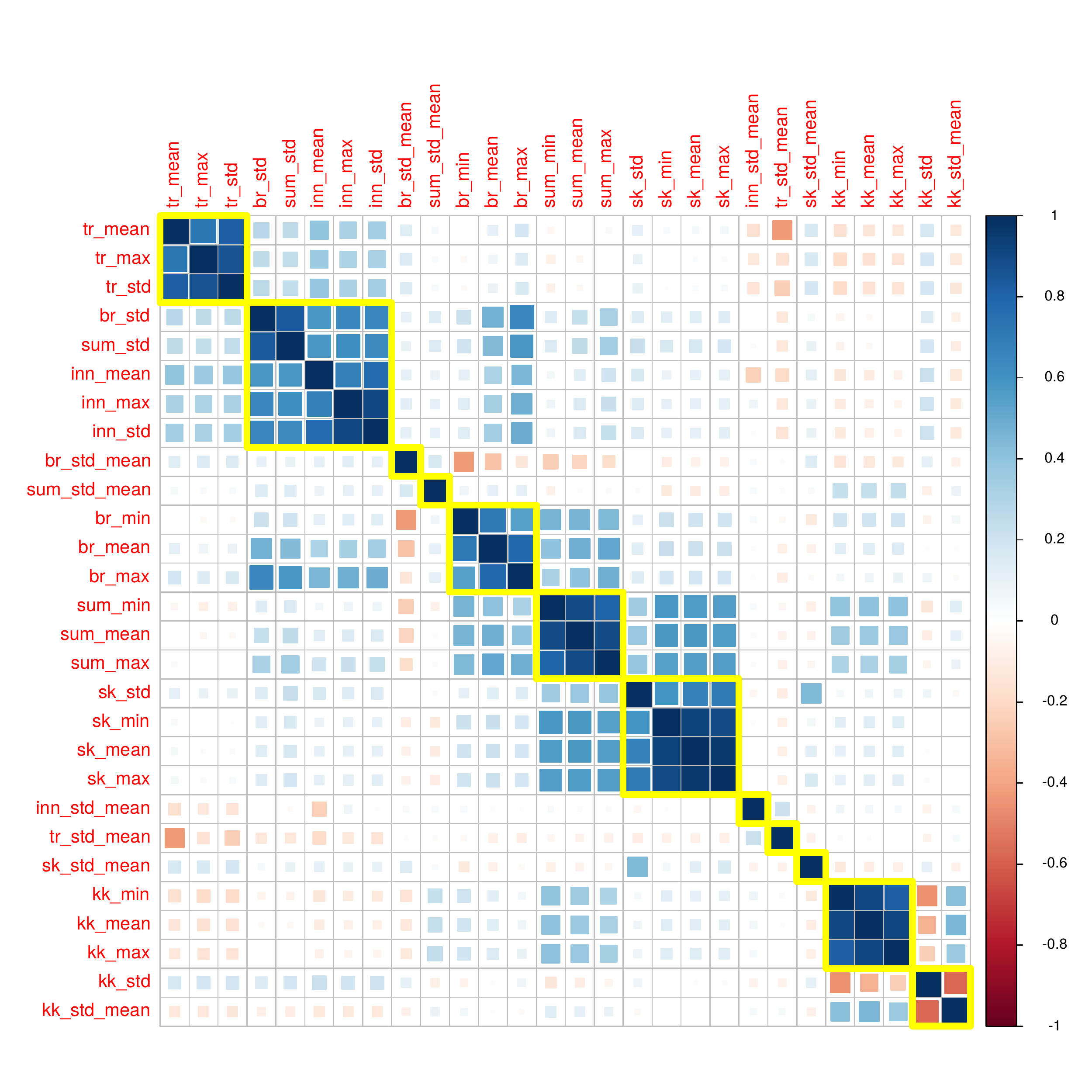}
	\end{center}
	\caption{Kendall's $\tau$ values for the real data set. The chosen clusters are marked with a yellow rectangle.
		\label{fig:dnb}}
\end{figure}

In our last example, we use a real data set. The data set consists of 28 features extracted from 6 transaction time series. It has previously been used for predicting mortgage default, 
relating probability of default to transaction information \citep{kvamme2018}. The transaction information consists of the daily balances on consumers' credit ({\tt kk}), checking ({\tt br}), and savings ({\tt sk}) accounts, 
in addition to the daily number of transactions on the 
checking account ({\tt tr}), the amount transferred into the checking account ({\tt inn}), and the sum of the checking, savings, and credit card accounts  ({\tt sum}).
For each of these time series,  which were of length 365 days, the mean ({\tt mean}), maximum ({\tt max}), minimum ({\tt min}), standard deviation ({\tt std}), and the standard deviation 
scaled by the mean ({\tt std\_mean}) were computed, resulting in 28 features. These are scaled to have mean 0 and standard deviation 1 before they are used in our computations. Figure \ref{fig:histReal} shows histograms 
for nine of the features (the remaining features have similar distributions). The feature distributions are skewed and heavy-tailed. The pairwise rank correlations (measured by Kendall's $\tau$, see Section
\ref{KendTau} for more information) are shown in Figure \ref{fig:dnb}. Most correlations are close to 0, but as can be seen from the figure, there are groups of features with high mutual correlations.   

The data set was divided into a training set and a test set, containing 12,696 and 1,921 observations respectively. 
We fitted an XGBoost model with 50 trees to the training data, using default parameter settings. The resulting AUC for the test data was 0.88. 

\begin{figure}[ht]
	\begin{center}
		\includegraphics[width=0.7\linewidth]{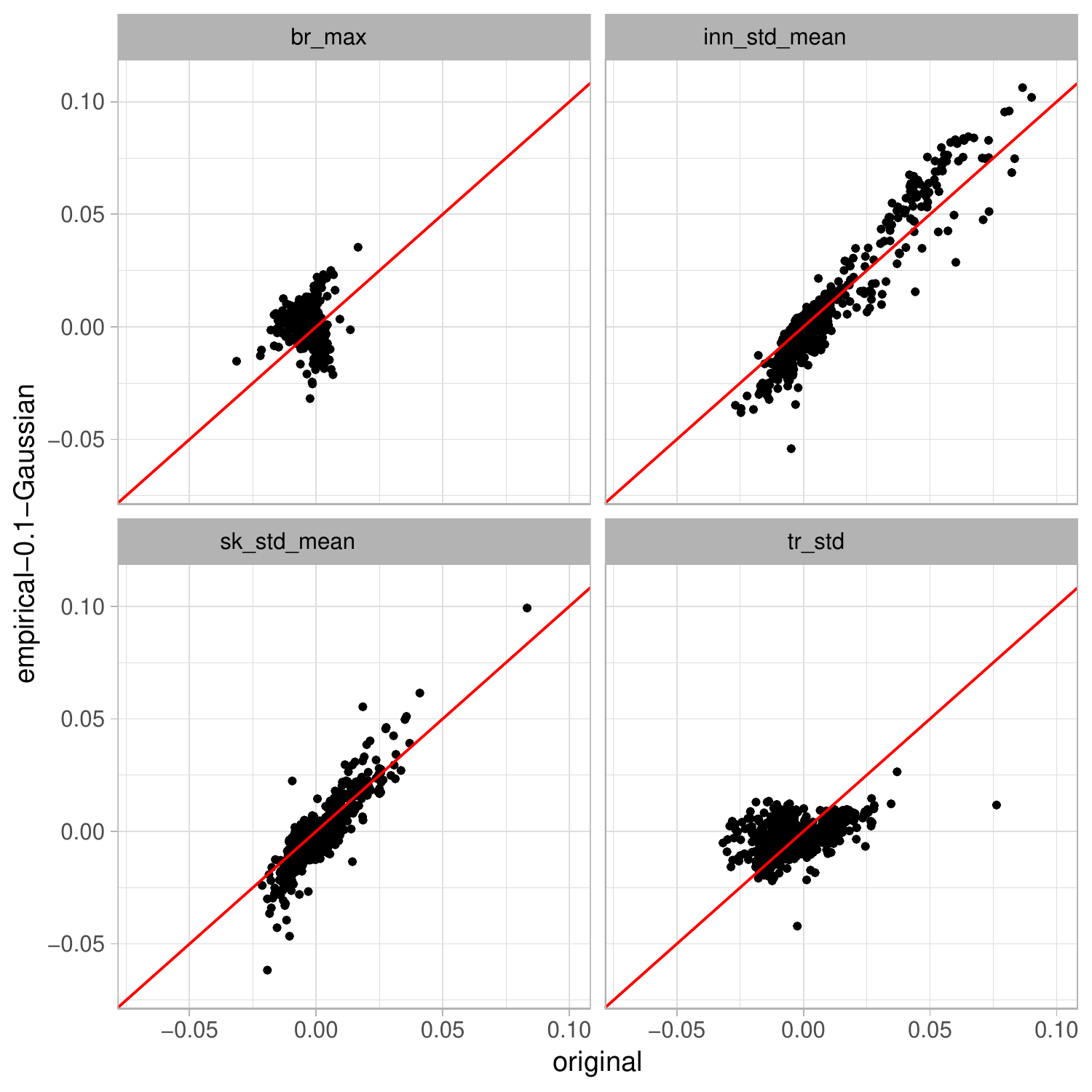}
	\end{center}
	\caption{Shapley values estimated with the original Kernel SHAP approach against those estimated with the empirical-0.1+Gaussian method for four of the features.
		\label{corPlots}}
\end{figure}

The last example in Section \ref{dim10} showed that for the case with skewed and heavy-tailed features and a piecewise constant model, the combined approaches were superior to the other, with the empirical-0.1+Gaussian 
approach as the best performing method. We assume that this is also the case for the real data set, and compare the performance of this method with the original Kernel SHAP approach. Figure \ref{corPlots} shows plots of 
Shapley values estimated with the original Kernel SHAP approach against those estimated with the empirical-0.1+Gaussian method for four of the features. For the features \texttt{inn\_std\_mean} and \texttt{sk\_std\_mean}
there is a high degree of correspondence, while for \texttt{tr\_std} and \texttt{br\_max}, the Shapley values for the two methods are very different. These results may at least partly be explained by 
Figure \ref{fig:dnb}. The features \texttt{inn\_std\_mean} and \texttt{sk\_std\_mean} are almost independent from all other features, while \texttt{tr\_std}, and especially \texttt{br\_max}, have a high degree of dependence with many
other features. Hence, the original Kernel SHAP method is likely to perform well for \texttt{inn\_std\_mean} and \texttt{sk\_std\_mean} and poor for \texttt{tr\_std} and \texttt{br\_max}.


\section{Clustering of Shapley values corresponding to feature dependence\label{sec:clustering}}

Using any of our suggested Shapley value approximations, the
dependencies between features are handled in a proper way. One way to
then visually present the explanation of a particular prediction could
be to rank the absolute Shapley values and present them and their
corresponding features in descending order, for example for the ten
most important features. However, if some or many of the features are
dependent, a grouping or clustering of dependent features may ease
the interpretation and use of the Shapley values. 


Assume that we have many features that are highly correlated, but none of them is among the ten most important features. If presented as a group, however, their combined Shapley value might turn out to be on the top ten list. Even when the individual Shapley values are well approximated, clustering can therefore aid in presenting the model explanation more correctly.

This clustering of features can in principle be different for each individual, but that would make model explanations across individuals opaque. We therefore suggest to cluster the features once on 1) the training data, or 2) test data, given that the test data set is large enough for reliable clustering to be done. We are then left with the choice of dependence measure and  clustering method. Our preferred choices are not canonical in any sense and other choices may suit your particular application better.

\subsection{Dependence measure}\label{KendTau}
The Pearson correlation coefficient measures the linear
correlation. As our Shapley values are not restricted to linear
dependence, we suggest to apply a more robust measure of dependence
with fewer implicit assumptions, namely the rank correlation, more
specifically as measured by Kendall's $\tau$
\citep{kendall38}. Kendall's $\tau$ for two random features $x_j$ and $x_k$ is defined as
\begin{align*}
T_{j,k} = \frac{1}{n(n-1)}\sum_{i \neq l} \sign(x_{j}^i-x_{j}^l)\sign(x_{k}^i-x_{k}^l),
\end{align*}
where $i,l=1,\ldots n$ are the samples we use to compute this measure.

\subsection{Clustering method and cluster size}

First, we define a dissimilarity  matrix, that is distances between pairs of features. Since we would like to group features together that are strongly correlated, regardless of whether this correlation is positive or negative, we define a 
dissimilarity matrix $\boldsymbol{D}$ having indicies
\begin{equation}\label{distanceFunc}
D_{j,k} = 1- |T_{j,k}|, \quad j,k = 1,\ldots,M,
\end{equation}
where $|(\cdot)|$ denotes the absolute value.
Hence, two features that are perfectly correlated will have a dissimilarity value of 0, while a pair of independent features will have a dissimilarity value of 1. We then do  hierarchical, agglomerative clustering with the agglomeration method commonly denoted 
as ``complete'' \citep{mullner13} and $\boldsymbol{D}$ in \eqref{distanceFunc} as dissimilarity matrix. This means that when merging of two clusters $J$ and $K$ is considered, the cluster dissimilarity is given as
\begin{align*}
d_{J,K} = \max_{j\in J,k\in K} D_{j,k}.
\end{align*}
Further, suppose now that $J$ and $K$ are clusters that are joined into a new cluster, and $C$ is a any other cluster. The distance between the joined cluster $J \cup K$ and cluster $C$ is then defined as
\begin{align*}
\max (d_{J,C},d_{K,C}).
\end{align*}

The number of clusters can be chosen either by a user chosen dissimilarity level or some sort of optimal choice of clusters. We used the minimum value of the Kelley-Gardner-Sutcliffe penalty function for a hierarchical cluster tree \citep{kelley96}. 
In the implementation of this method in \cite{maptree12}, there is an additional tuning parameter $\alpha$ with default value equal to 1 that can be used to scale the optimal number of clusters.

\subsection{Real data example continued}

\begin{figure}[htb!]
	\begin{center}
		\begin{subfigure}[t]{0.49\linewidth}
			\includegraphics[width=\linewidth]{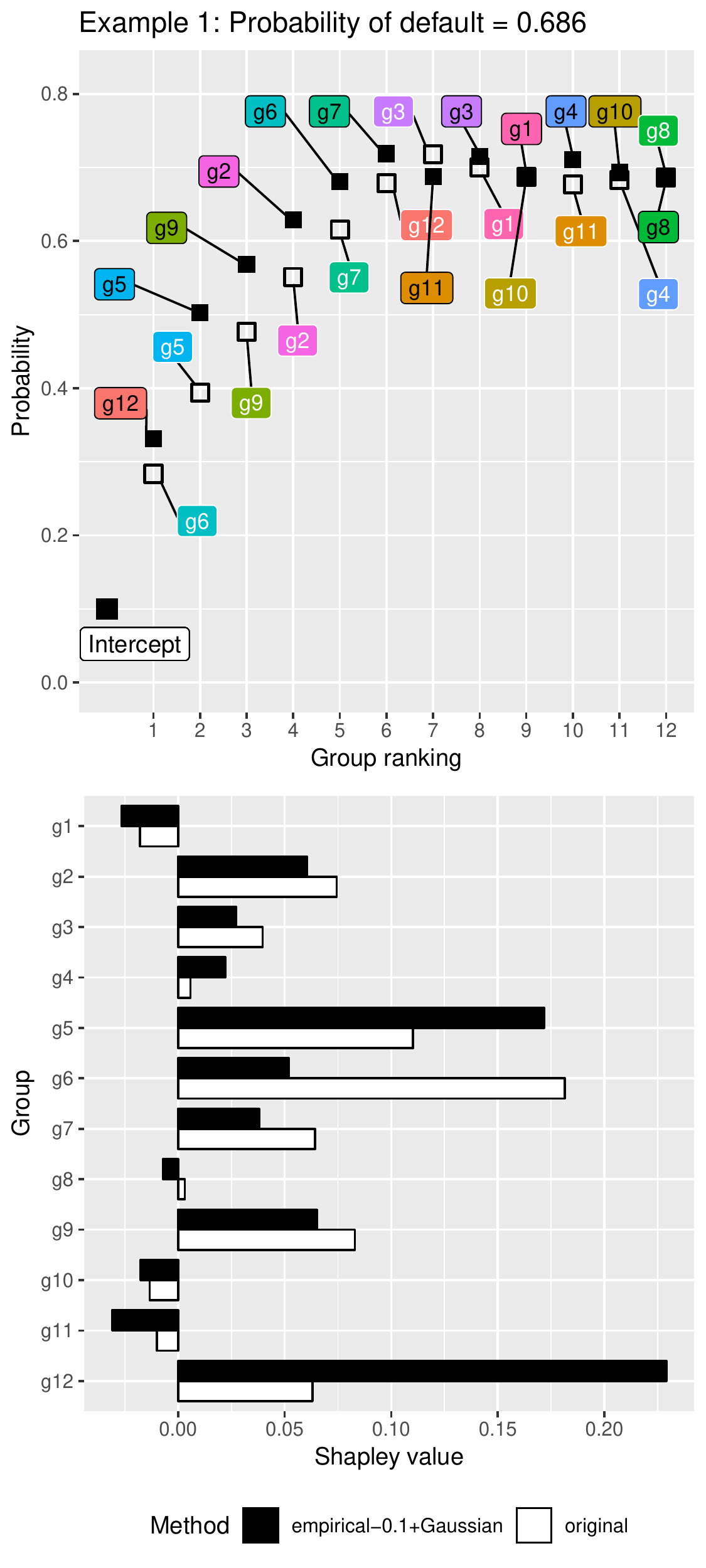} 
		\end{subfigure}
		\rulesep
		\begin{subfigure}[t]{0.49\linewidth}
			\includegraphics[width=\linewidth]{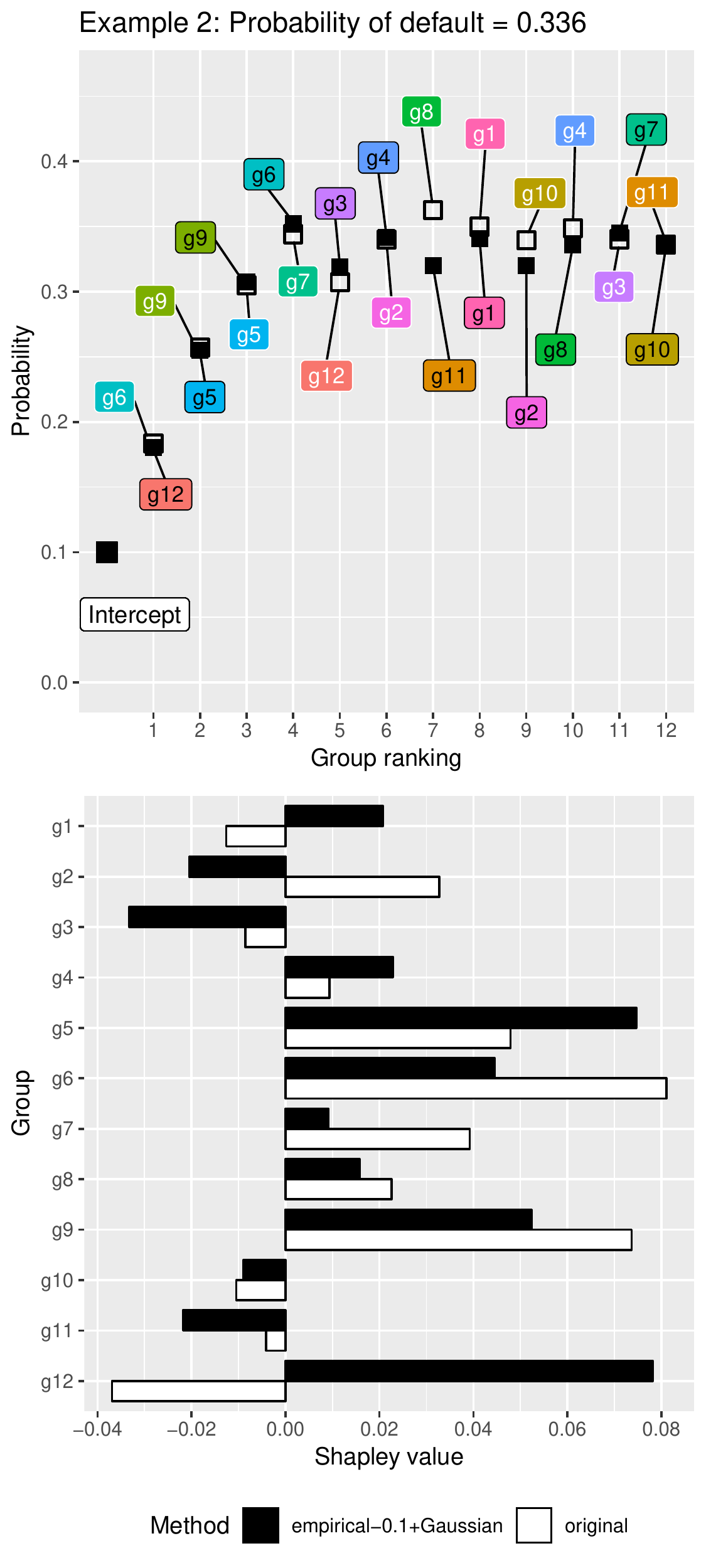} 
		\end{subfigure}
		\caption{The contribution to the probability of default from the twelve groups of features according to the clustering in Figure \ref{fig:dnb}. Two examples are shown with two methods. The top panels is a waterfall chart of the contributions, where the groups are ranked based on the absolute Shapley value. Starting with the intercept, the vertical step between every group is equal to the Shapley value of that group, such that the predicition is reached on the right hand side of the plot. The bottom panels show the Shapley values for each group.
			\label{fig:real_data_ex_comb}}
	\end{center}
\end{figure}

Here, we continue with the example from Section \ref{sec:real}. Using the Kelley-Gardner-Sutcliffe penalty function with $\alpha=0.1$, the features were clustered into twelve groups, denoted $\text{g1}, \text{g2}, \ldots, \text{g12}$  
from left to right in Figure \ref{fig:dnb}. Hence, g1 contains the features \texttt{tr\_mean}, \texttt{tr\_max} and \texttt{tr\_std} and g3 contains \texttt{br\_std\_mean} only, etc. Figure \ref{fig:real_data_ex_comb} shows the combined Shapley values 
for the twelve groups for two example predictions. In both examples it is apparent from the top panels that the original Kernel SHAP method has a ranking of groups that is very different from the empiricial-0.1+Gaussian method, which we believe is more correct.
Further, as indicated by the lower panels, the Shapley values obtained by the original Kernel SHAP method and the empiricial-0.1+Gaussian method are very different for the majority of the groups. Take for instance example 2. For this person, group 12 is
considered to be the one providing the highest \textit{increase} in probability by the empirical-0.1+Gaussian approach, while according to the original Kernel SHAP, group 12 is the one \textit{reducing} the probability the most. 
This means that the original Kernel SHAP approach give completeley different explanations from our approach, which is regarded to be more trustworthy according to our simulation experiments in Section \ref{experiment}.
In real data settings it is crucial to get a correct explanation, as an incorrect explanation may cause wrong conclusions and actions. Hence, it is very important to take the feature dependence into account.

%

\section{Summary and discussion}\label{summary}
\noindent 

Shapley values is a model-agnostic method for explaining individual predictions with a solid theoretical foundation. The main disadvantage with this method is that the computational time grows exponentially. 
This has led to approximations, of which the 
Kernel SHAP method is the most known. A key ingredient of the Kernel SHAP method is the conditional distribution of a subset $\bar{\mathcal{S}}$ of the features, conditional on the features in $\mathcal{S}$ that are not in this subset. It is assumed that the features 
in the two subsets are independent, meaning that the conditional distribution may be replaced by the marginal distribution of the features in $\bar{\mathcal{S}}$. If there is a high degree of dependence among some or all the features, the resulting 
explanations might be very wrong. This paper introduces a modified version of the Kernel SHAP method, which handles dependent features. We have proposed four different approaches for estimating the conditional distribution;
assuming a Gaussian multivariate distribution for all features, assuming a Gaussian copula with empirical margins, using an empirical approach and a combination of the empirical approach and either the Gaussian or the Gaussian copula approach. 

We have performed a comprehensive simulation study with linear and non-linear models, Gaussian and non-Gaussian distributions, and dimensions 3 and 10,  where our methods give more accurate approximations to the true Shapley values than the 
original Kernel SHAP approach. For the non-linear models, these methods clearly outperformed the Tree SHAP method, which, to the best of our knowledge, is the only Shapley  approach 
which tries to handle dependence between features in the prediction explanation setting. When performing our experiments, it turned out that the non-parametric approach was superior when conditioning on a small number of the features, 
while it was outperformed by the Gaussian and copula methods if we condition on more variables. Hence, we regard the combined approach to be the most promising of our proposed methods. This approach was applied to a real case with 28 variables, where the 
predictions to be explained were produced by a XGBoost classifier designed to predict mortgage default. In this case, the true Shapley values are not known.  However, we provide results, which indicate that our combined approach provides 
more sensible approximations than the Tree SHAP and original Kernel SHAP methods.         

If features are dependent, it is more challenging to interpret the resulting Shapley values. We have therefore proposed a method for aggregating individual Shapley values. Using hierarchical agglomerative 
clustering of the training data, the features are divided into a number of clusters. Having obtained the individual Kernel SHAP values using one of the approaches above, we compute the sum of these values for 
each cluster and visually present these sums instead of the individual values.           

While having many desirable properties, our method has one obvious drawback: the computational time. The most time-consuming part is the AICc computation of the bandwidth parameter in the non-parametric approach. Using a fixed
bandwidth parameter instead, the computational time of the combined approach is the same as that of the original Kernel SHAP method. Very recently, there has been some attempts at using the 
underlying graph structure of the input data to reduce the computational complexity \citep{Chen2018,Li2018}. If several conditional independence requirements are satisfied, the full graph may first be divided into separate communities and then the 
Kernel SHAP method may be applied to each community individually. The methods proposed by \cite{Chen2018,Li2018} were tested on predictions obtained from text and image classification. In such settings the data often has a graph structure, 
enabeling factorization into separate communities. The main challenge with using this method for tabular data is the potentially huge number of tests for conditional dependence which has to be performed. However, it is definitely worth a closer look. 

In this paper, we assume that the aim is to explain the actual predictions from the model. For some models, the prediction is a probability. If the Shapley framework is used to decompose probabilities, summing over a subset of the $\phi_i$-values 
is likely to produce values that are not in the range [0,1]. Hence, in such cases it might be more natural to assume that the importance of features is additive in the log odds space rather than in the space of probabilities. There are however 
problems even with this solution, since it is not straightforward for a human to interpret a log odds contribution to the probability. Hence, what to decompose seems to be more a practical than a mathematical question.

Tabular data often contain ordered or even non-ordered categorical data. The proposed non-parametric approach may still be used if the categorical variables are converted into numerical ones. The 
simplest solution is to use one-hot-encoding. However, large data sets often handle categorical features with hundreds of categories, meaning that 
a such method needs to handle a large number of binary attributes. An alternative approach is to use ideas from the clustering literature \citep{Huang97, Huang98} defining distance functions that handle both categorical and mixed-type 
features. There are also generalizations of the Mahalanobis distance treating data with a mixture of nominal, ordinal and continuous variables that might be used instead of the approach described above, see for instance \cite{DELEON05}. 
When it comes to the parametric approaches, categorical data represents a greater challenge. The most promising alternative might be to use entity embedding \citep{Guo16} to convert the categorical features into numerical 
ones, and then treat these features similarly to the other numerical features.


\section*{Acknowledgement}
The authors are grateful to Scott Lundberg  for valuable advice concerning the Kernel SHAP method. We also want to thank Ingrid Hob\ae k Haff for suggesting the copula approach, Nikolai Sellereite for major contributions to the implementation of the methods in
the R-package \verb,shapr, (\url{https://github.com/NorskRegnesentral/shapr}). This work was supported by the Norwegian Research Council grant 237718 (Big Insight).

\appendix

\section{Shapley properties in the prediction explanation setting}\label{properties}
When using Shapley values for prediction explanation, $\phi_0 = v(\emptyset)=\text{E}[f(\boldsymbol{x})]$ actually plays an important role, quantifying how much of a prediction which is not due to \textit{any} of the features, but merely to the global average prediction. If $\phi_0$ is large (in absolute value) compared to $\phi_j, j\neq0$, the features are said to be 'not important' for that specific prediction. Moreover, in the prediction explanation setting, the interpretation of the properties of the Shapley value discussed in Section \ref{ExactShap} are as follows:
\begin{description}
	\item[Efficiency:] The sum of the Shapley values for the different features is equal to the difference between the prediction and the global average prediction: $\sum_{j=1}^M \phi_j  = f(\boldsymbol{x}^*) -\text{E}[f(\boldsymbol{x})]$. This property ensures that
	the part of the prediction value which is not explained by the global mean prediction, is fully devoted and explained by the features, and that the $\phi_j, j=1,\ldots,M$ can be compared across different predictions $f(\boldsymbol{x}^*)$.
	\item[Symmetry:] The Shapley values for two features are equal if, when combined with any other subsets of features, they contribute equally to the prediction. Failure to fulfill such a criterion, that is, that the same contribution from two different features does 
	not give the same explanation, would be inconsistent and give untrustworthy explanations.
	\item[Dummy player:] A feature that does not change the prediction, no matter which other features it is combined with, has a Shapley value of 0. Assigning a nonzero explanation value to a feature that has no influence on the prediction would be very odd, so this is a natural requirement.
	\item[Linearity:] When a prediction function consists of a sum prediction functions, the Shapley value for a feature is identical to the sum of that feature's Shapley values from each of the individual prediction functions. This also holds for linear combinations of prediction 
	functions. This property ensures that models on this form, such as Random Forest or other structurally simple ensemble models, can be explained and interpreted individually.
\end{description}
Failure to fulfill any of these basic and advantageous properties gives an odd, undesirable or inconsistent explanation framework. There is no other additive explanation method than Shapley values which satisfy all these criteria \citep{Lundberg}.



\section{Shapley values when the model is linear}\label{appLinear}

In this section we first give a proof for the explicit formula for the Shapley values when the predictive model is a linear regression model, and all features are independent. Then, we show how to obtain the contribution 
function $v(\mathcal{S})$ when the model still is linear, but the features might be dependent.

\subsection{Linear model and independent features}

When $v(\mathcal{S}) = \mbox{E}[f(\boldsymbol{x})|\boldsymbol{x}_{\mathcal{S}}=\boldsymbol{x}_{\mathcal{S}}^*]$, the predictive model is a linear regression model, and all features are independent, then the Shapley values 
take the simple form
\[  \phi_{j} = \beta_j\,(x^*_{j}-E[x_j]), \quad j = 1,\ldots,M.\]

\paragraph{Proof:} First, we derive the expression for $v(\mathcal{S})$ in this case:

\begin{align}
v(\mathcal{S}) &= E[f(\boldsymbol{x})|\boldsymbol{x}_{\mathcal{S}}=\boldsymbol{x}_{\mathcal{S}}^*]\nonumber\\
&= \int f(\boldsymbol{x}_{\bar{\mathcal{S}}},\boldsymbol{x}_{\mathcal{S}}^*)\,p(\boldsymbol{x}_{\bar{\mathcal{S}}}|\boldsymbol{x}_{\mathcal{S}}=\boldsymbol{x}_{\mathcal{S}}^*)\,d\boldsymbol{x}_{\bar{\mathcal{S}}}, \label{step1} \\
&= \int\left(\sum_{i \in \bar{\mathcal{S}}} \beta_i\,x_i + \sum_{i \in \mathcal{S}} \beta_i\,x_i^*\right)\,p(\boldsymbol{x}_{\bar{\mathcal{S}}}|\boldsymbol{x}_{\mathcal{S}}=\boldsymbol{x}_{\mathcal{S}}^*)\,d\boldsymbol{x}_{\bar{\mathcal{S}}} \label{step2}\\
&=  \sum_{i \in \bar{\mathcal{S}}}\beta_i \int x_i\,\,p(x_i)\,dx_i + \sum_{i \in \mathcal{S}}\beta_i\,x_i^*\int p(\boldsymbol{x}_{\bar{\mathcal{S}}})\,d\boldsymbol{x}_{\bar{\mathcal{S}}} \label{step3}\\
&=  \sum_{i \in \bar{\mathcal{S}}}\beta_i\,E[x_i] + \sum_{i \in \mathcal{S}}\beta_i\,x_i^* \nonumber
\end{align}
The transition from step \eqref{step1} to \eqref{step2} follows from the assumption of a linear model, while the transition from \eqref{step2} to \eqref{step3} follows from the independent features assumption. 

Having computed $v(\mathcal{S})$, the expression for $v(\mathcal{S} \cup \{j\})$ can be simply found as
\[ v(\mathcal{S} \cup \{j\})=v(\mathcal{S}) + \beta_j\,x_j^* -\beta_j\,E[x_j],\]
meaning that 
\[v(\mathcal{S} \cup \{j\})-v(\mathcal{S}) = \beta_j\left(x_j^* -E[x_j]\right).\]
From the above, we see that in this case, the difference $v(\mathcal{S} \cup \{j\})-v(\mathcal{S})$ is independent of $\mathcal{S}$. Hence, the Shapley formula may be written as
\[\phi_j = \beta_j\left(x_j^* -E[x_j]\right)\sum_{\mathcal{S} \subseteq \mathcal{M} \setminus\{j\}} \frac{|\mathcal{S}| ! (M-| \mathcal{S}| - 1)!}{M!}. \]
Since the sum of the Shapley weights is 1, we have
\[\phi_j = \beta_j\left(x_j^* -E[x_j]\right),\]
for $j=1,\ldots,M. \blacksquare$

\subsection{Linear model and dependent features}

If the model is linear, but the features are {\it not} independent, $v(\mathcal{S})$ instead may be derived as follows
\begin{align}
v(\mathcal{S}) &= E[f(\boldsymbol{x})|\boldsymbol{x}_{\mathcal{S}}=\boldsymbol{x}_{\mathcal{S}}^*]\nonumber\\
&= \int f(\boldsymbol{x}_{\bar{\mathcal{S}}},\boldsymbol{x}_{\mathcal{S}}^*)\,p(\boldsymbol{x}_{\bar{\mathcal{S}}}|\boldsymbol{x}_{\mathcal{S}}=\boldsymbol{x}_{\mathcal{S}}^*)\,d\boldsymbol{x}_{\bar{\mathcal{S}}}, \label{step1b} \\
&= \int\left(\sum_{i \in \bar{\mathcal{S}}} \beta_i\,x_i + \sum_{i \in \mathcal{S}} \beta_i\,x_i^*\right)\,p(\boldsymbol{x}_{\bar{\mathcal{S}}}|\boldsymbol{x}_{\mathcal{S}}=\boldsymbol{x}_{\mathcal{S}}^*)\,d\boldsymbol{x}_{\bar{\mathcal{S}}} \label{step2b}\\
&=  \sum_{i \in \bar{\mathcal{S}}}\beta_i \int x_i\,\,p(x_i| \boldsymbol{x}_{\mathcal{S}}=\boldsymbol{x}_{\mathcal{S}}^*)\,dx_i + \sum_{i \in \mathcal{S}}\beta_i\,x_i^*\int p(\boldsymbol{x}_{\bar{\mathcal{S}}})\,d\boldsymbol{x}_{\bar{\mathcal{S}}} \label{step3b}\\
&=  \sum_{i \in \bar{\mathcal{S}}}\beta_i\,E[x_i|\boldsymbol{x}_{\mathcal{S}}=\boldsymbol{x}_{\mathcal{S}}^*] + \sum_{i \in \mathcal{S}}\beta_i\,x_i^* \nonumber \\
&= f(\boldsymbol{x}_{\bar{\mathcal{S}}}=E[\boldsymbol{x}_{\bar{\mathcal{S}}}|\boldsymbol{x}_{\mathcal{S}}=\boldsymbol{x}_{\mathcal{S}}^*], \boldsymbol{x}_{\mathcal{S}}=\boldsymbol{x}_{\mathcal{S}}^*)
\end{align}
In this case, one may therefore avoid time-consuming simulations if one is able to analytically obtain a proper estimate of $E[\boldsymbol{x}_{\bar{\mathcal{S}}}|\boldsymbol{x}_{\mathcal{S}}=\boldsymbol{x}_{\mathcal{S}}^*]$. This
is not straightforward in general.

\section{Generalized Hyperbolic Distribution}\label{app1}

The density of a $d$-dimensional Generalized Hyperbolic random vector $\boldsymbol{X}$ is
\[ p(\boldsymbol{x}) = \left[\frac{\omega + \delta(\boldsymbol{x}, \boldsymbol{\mu},\boldsymbol{\Sigma})}{\omega + \boldsymbol{\beta}^T\boldsymbol{\Sigma}^{-1}\boldsymbol{\beta}}\right]^{\frac{\lambda-d/2}{2}}
\frac{K_{\lambda-d/2}\left(\sqrt{(\omega + \delta(\boldsymbol{x}, \boldsymbol{\mu},\boldsymbol{\Sigma}))(\omega + \boldsymbol{\beta}^T\boldsymbol{\Sigma}^{-1}\boldsymbol{\beta})}\right)}
{(2\pi)^{d/2}|\boldsymbol{\Sigma}|^{1/2}K_{\lambda}(\omega)\exp\left\{-(\boldsymbol{x}-\boldsymbol{\mu})^T\boldsymbol{\Sigma}^{-1}\boldsymbol{\beta}\right\}},\]
where $\delta(\boldsymbol{x}, \boldsymbol{\mu},\boldsymbol{\Sigma})=(\boldsymbol{x}-\boldsymbol{\mu})^T\boldsymbol{\Sigma}^{-1}(\boldsymbol{x}-\boldsymbol{\mu})$ is the squared Mahalanobis distance between
$\boldsymbol{x}$ and $\boldsymbol{\mu}$, and $K_{\lambda}$ is the modified Bessel function of the third kind with index $\lambda$. The mean vector and covariance matrix of $\boldsymbol{X}$ are
\[ \mbox{E}(\boldsymbol{X})=\boldsymbol{\mu} + \mbox{E}(W)\boldsymbol{\beta} \hspace{1cm} \mbox{and} \hspace{1cm} \mbox{Var}(\boldsymbol{X})=\mbox{E}(W)\boldsymbol{\Sigma} + \mbox{Var}(W)\boldsymbol{\beta}\boldsymbol{\beta}^T.\]

It can be shown \citep{WEI201918} that if $\boldsymbol{X}$ is partitioned as $(\boldsymbol{X}_1^T,\boldsymbol{X}_2^T)^T$, where $\boldsymbol{X}_1$ is $d_1$-dimensional and $\boldsymbol{X}_2$ is $d_2$-dimensional,
the conditional distribution of $\boldsymbol{X}_2$ given that $\boldsymbol{X}_1=\boldsymbol{x}_1$ is Generalized Hyperbolic distributed, that is, 
$\boldsymbol{X}_2|\boldsymbol{X}_1=\boldsymbol{x}_1 \sim \mbox{GH}^*(\lambda_{2|1},\chi_{2|1},\psi_{2|1},\boldsymbol{\mu}_{2|1},\boldsymbol{\Sigma}_{2|1},\boldsymbol{\beta}_{2|1})$, where
\begin{alignat*}{3}
\lambda_{2|1} &= \lambda - d_1/2, 
\quad\quad\quad\quad\quad\quad\quad\quad\quad\quad\quad \boldsymbol{\mu}_{2|1} && =  \boldsymbol{\mu}_{2} + \boldsymbol{\Sigma}_{12}^T\boldsymbol{\Sigma}_{11}^{-1}(\boldsymbol{x}_1-\boldsymbol{\mu}_1),\\
\chi_{2|1}&= \omega + (\boldsymbol{x}_1-\boldsymbol{\mu}_1)^T\boldsymbol{\Sigma}_{11}^{-1}(\boldsymbol{x}_1-\boldsymbol{\mu}_1), \quad
\quad \boldsymbol{\Sigma}_{2|1} && =  \boldsymbol{\Sigma}_{22} - \boldsymbol{\Sigma}_{12}^T\boldsymbol{\Sigma}_{11}^{-1}\boldsymbol{\Sigma}_{12},\\
\psi_{2|1} &= \omega + \boldsymbol{\beta}_1^T\Sigma_{11}^T\boldsymbol{\beta}_1, \quad\quad\quad\quad\quad\quad\quad\quad\quad
\boldsymbol{\beta}_{2|1} && =   \boldsymbol{\beta}_{2} - \boldsymbol{\Sigma}_{12}^T \boldsymbol{\Sigma}_{11}^{-1}\boldsymbol{\beta}_1.
\end{alignat*}

Note that GH* means that a slightly different parameterization of the GH distribution is used for the conditional distribution. Here, due to technical reasons, we use the parameterization proposed by \cite{McNeilBook}:
\[ p(\boldsymbol{x}) = \left[\frac{\chi + \delta(\boldsymbol{x}, \boldsymbol{\mu},\boldsymbol{\Sigma})}{\psi + \boldsymbol{\beta}^T\boldsymbol{\Sigma}^{-1}\boldsymbol{\beta}}\right]^{\frac{\lambda-d/2}{2}}
\frac{(\psi/\chi)^{\lambda/2}K_{\lambda-d/2}\left(\sqrt{(\chi + \delta(\boldsymbol{x}, \boldsymbol{\mu},\boldsymbol{\Sigma}))(\psi + \boldsymbol{\beta}^T\boldsymbol{\Sigma}^{-1}\boldsymbol{\beta})}\right)}
{(2\pi)^{d/2}|\boldsymbol{\Sigma}|^{1/2}K_{\lambda}(\sqrt{\chi\psi})\exp\left\{-(\boldsymbol{x}-\boldsymbol{\mu})^T\boldsymbol{\Sigma}^{-1}\boldsymbol{\beta}\right\}}.\]



\vskip 0.2in
\bibliographystyle{biometrika_new}
\bibliography{references}

\end{document}